\documentclass{article}

\usepackage{PRIMEarxiv}

\usepackage{amsfonts}       
\usepackage{amsmath}
\usepackage{amssymb}
\usepackage{amsthm}
\usepackage{booktabs} 
\usepackage{cancel}
\usepackage{caption}
\usepackage{changepage}
\usepackage{enumitem}
\usepackage{fancyhdr}       
\usepackage{float}  
\usepackage[T1]{fontenc}    
\usepackage{graphicx}       
\usepackage{hyperref}       
\usepackage[utf8]{inputenc} 
\usepackage{lipsum}
\usepackage{mathtools}
\usepackage{microtype}      
\usepackage{multicol, multirow} 
\usepackage{natbib}
\usepackage{nccmath}
\usepackage{nicefrac}       
\usepackage{setspace}
\usepackage{subfigure}

\usepackage{url}            

\usepackage[utf8]{inputenc} 
\usepackage[T1]{fontenc}    
\usepackage{hyperref}       
\usepackage{url}            
\usepackage{booktabs}       
\usepackage{amsfonts}       
\usepackage{nicefrac}       
\usepackage{microtype}      
\usepackage{lipsum}
\usepackage{fancyhdr}       
\usepackage{graphicx}       
\usepackage{tikz, pgfplots}
\graphicspath{{media/}}     
\tikzset{>=latex}

\usepackage{todonotes}

\usepackage[utf8]{inputenc} 
\usepackage[T1]{fontenc}    
\usepackage{hyperref}       
\usepackage{url}            
\usepackage{booktabs}       
\usepackage{amsfonts}       
\usepackage{nicefrac}       
\usepackage{microtype}      
\usepackage{xcolor}         

\graphicspath{{media/}}     

\title{Graph Machine Learning based Doubly Robust Estimator for Network Causal Effects
}

\author{
S. Baharan Khatami\\
  UC San Diego\\
  \texttt{skhatami@ucsd.edu}
  \And 
    Harsh Parikh\\
  Johns Hopkins University\\
  \texttt{hparikh4@jh.edu}
    \And
  Haowei Chen\\
  UC San Diego\\
  \texttt{hac034@ucsd.edu}
  \And 
  Sudeepa Roy\\
  Duke University\\
  \texttt{sudeepa@cs.duke.edu}
   \And
  Babak Salimi \\
  UC San Diego \\
  \texttt{bsalimi@ucsd.edu}
}

\theoremstyle{plain}
\newtheorem{theorem}{Theorem}[section]
\newtheorem{proposition}[theorem]{Proposition}

\theoremstyle{definition}
\newtheorem{definition}[theorem]{Definition}
\newtheorem{assumption}[theorem]{Assumption}
\theoremstyle{remark}

\usepackage{times}
\usepackage{helvet}
\usepackage{courier}

\usepackage{tikz}
\usetikzlibrary{positioning, arrows.meta, bending}
\usepackage[utf8]{inputenc}
\usepackage[utf8]{inputenc}
\usepackage{amsthm}
\usepackage{amssymb}
\usepackage{pifont}
\usepackage{dsfont}
\usepackage{color}
\usepackage{enumitem}
\usepackage{ulem}
\usepackage{float}
\usepackage{lscape}
\usepackage{rotating}
\usepackage{longtable}

\usepackage{etoc}

\usepackage{mdframed}

\newcommand{\vect}[1]{\mathbf{#1}} 
\newcommand{\matr}[1]{\mathcal{#1}} 
\newcommand{\set}[1]{\mathbb{#1}} 

\newcommand{\doubleprime}{{\prime\prime}}

\usepackage{ctable}

\definecolor{amber}{rgb}{1.0, 0.75, 0.0}

\begin{document}
\maketitle

\begin{abstract}
We address the challenge of inferring causal effects in social network data. This results in challenges due to interference --where a unit's outcome is affected by neighbors' treatments -- and network-induced confounding factors.
While there is extensive literature focusing on estimating causal effects in social network setups, a majority of them make prior assumptions about the form of network-induced confounding mechanisms. Such strong assumptions are rarely likely to hold especially in high-dimensional networks. We propose a novel methodology that combines graph machine learning approaches with the double machine learning framework to enable accurate and efficient estimation of direct and peer effects using a single observational social network. We demonstrate the semiparametric efficiency of our proposed estimator under mild regularity conditions, allowing for consistent uncertainty quantification. We demonstrate that our method is accurate, robust, and scalable via extensive simulation study. We use our method to investigate the impact of Self-Help Group participation on financial risk tolerance. 
\end{abstract}

\keywords{Causal Inference \and Semi-Parametric Inference \and Double Machine Learning \and Graph Neural Networks}

\section{Introduction}

Our paper addresses the challenge of causal inference from social networks, a problem crucial for decision-making across various vital domains such as social media, healthcare, and economic networks \citep{jackson2008social,ogburn2022causal, atanasov2016shock, gassen2014causal}. For instance, causal inference helps understand the impact of recommendation algorithms on user engagement and preferences in social media, evaluate the effect of public health interventions like self-quarantine or school closures on the spread of infectious diseases within specific communities, and assess the influence of participation in self-help groups (SHG) on individuals' financial behaviors. Our paper uses the latter as a case study to investigate whether participation in SHG affects financial risk-taking behavior, observable through outstanding loans as a proxy measure.

Estimating causal effects from observational network data is challenging due to several factors: First, dependencies among individuals deviate from the traditional i.i.d. assumption, causing standard methods to fail. Second, network dependencies introduce interference between units, where their neighbors' treatments influence an individual's outcome, further complicating causal analysis \citep{Hudgens2008TowardCI, ogburn2014causal, aronow2017estimating, HalloranHudgens2012, VanderWeeleTchetgen2011}. Third, a unit's neighbors' covariates act as confounders that require adjusting for a complex set of potentially high-dimensional covariates with variable sizes for each unit, influenced by network structure and topology \citep{vanderweele2013social, ogburn2014causal}. 

Existing techniques often use simple predefined aggregates to summarize network information and develop estimators with theoretical guarantees, relying on the assumption of sufficiency of these aggregates \citep{forastiere2021identification, salimi2020causal, ogburn2022causal, 8631471}, which can lead to a loss of critical network information and erroneous conclusions. Recent techniques using graph machine learning, such as Graph Neural Networks (GNNs), have been introduced to deal with high-dimensional network covariate information  \citep{Ma2020CausalIU, jiang2022estimating, Guo2020IGNITEAM}. However, these methods lack theoretical guarantees and do not offer valid confidence intervals in causal effect estimation—each of which is critical. While graph machine learning methods are powerful in modeling high-dimensional covariates, their application for causal inference is challenging due to regularization-induced bias, overfitting, and slower convergence rates \citep{chernozhukov2018double}. Our method provides a principled approach that can employ graph machine learning for causal inference with theoretical guarantees by addressing these challenges.

In this paper, we extend the Double Machine Learning (DML) framework \citep{chernozhukov2018double} to develop a regular and asymptotically linear estimator for causal effects for network data that can work with the latest graph ML methods to address high-dimensional network covariates effectively with provable guarantees. Our estimator employs graph ML algorithms, such as Graph Neural Networks (GNNs), to estimate two predictive models: the propensity score, which is the probability of treatment assignment given network covariates, and the outcome model, which predicts the outcome given network covariates. By leveraging ``double'' ML or ``orthogonalized'' ML and sample splitting, we can construct high-quality point and interval estimates of causal parameters, effectively addressing issues like slow convergence rates, model misspecification, and regularization bias.  Furthermore, our estimator is doubly robust, meaning it remains consistent if either the propensity score or the outcome model is correctly specified. We prove our estimator's consistency and asymptotic normality under specific assumptions, allowing for the establishment of valid confidence intervals, which is crucial for the practical application of causal inference methods in social networks.

\paragraph{Contributions.} The key contributions of this research include:

\begin{enumerate}[leftmargin=0.5cm, itemsep=0pt]
    \item We develop a method for causal inference from social network data that integrates graph machine learning techniques, leveraging ``orthogonalized'' ML and sample splitting to construct high-quality point and interval estimates, addressing issues like slow convergence rates, model misspecification, and regularization bias.
    \item We demonstrate the theoretical properties of our estimator, including consistency and asymptotic normality under certain assumptions, which enable the construction of valid confidence intervals.
    \item We evaluate our framework on three semi-synthetic datasets and compare its performance against six leading methods, demonstrating superior performance.
    \item We conduct a case study on real-world data to examine the impact of Self-Help Group participation on financial risk tolerance, showcasing the practical applicability of our approach.
\end{enumerate}

 In section 2, we discuss the related work and literature, setting the stage for our contribution within the existing body of knowledge. In section 3, we discuss the required notations and the setup of our problem, the causal estimands of interest, necessary assumptions, and the identifiability of our estimands. Section 4 unveils our proposed estimator, detailing the intricacies of our methodology. In Section 5, we present the theoretical results underpinning our estimator. Section 6 showcases the experimental and empirical results that validate our approach. Finally, Section 7 encapsulates our findings, reflecting on the broader implications and potential avenues for future research in this domain.

\section{Related Works}
In addressing the interference issue in causal inference, numerous strategies have been proposed, primarily focusing on two main categories: design strategies for experimental setups and inference strategies utilizing observational data for post-experiment analysis.\\
Design strategies involve the integration of network information and control over treatment assignments to mitigate interference and enhance the estimation causal effect. A key approach within this domain is cluster-based randomization\citep{bland_cluster_2004}, where it is assumed that interference occurs only within clusters and not between them, a concept known as partial interference~\citep{sobel}. This approach has been extensively developed, with variations like multilevel or two-stage randomization ~\citep{hudgens}, where treatment or control assignments are made initially at the group level and subsequently at the unit level within each group. Recent advancements continue to build upon these foundations ~\citep{baird2014designing, imai_causal_2021}.
As an extension of cluster-based randomization in networks, another popular design to mitigate interference in the network is graph cluster randomization ~\citep{Ugander2013}. Related studies are developed by ~\citep{eckles2016design, UganderYin+2023, pouget2019testing, karrer2021network}.\\
Concurrently, increasing attention has been paid to observational studies under network interference, most of which primarily depend on the partial interference assumption. Methods developed by ~\citep{Hudgens2008TowardCI, forastiere2021identification, tchetgen} are extensions of Inverse probability weighting (IPW) estimator to estimate treatment effects in the presence of interference. ~\citep{liuIPW} have further expanded on this, adapting the IPW estimators to more complex scenarios of interference within networks.
Contrasting these developments, our research introduces a doubly robust estimator. This estimator stands out for its enhanced efficiency compared to IPW estimators, even when only the treatment nuisance model is correctly specified.
~\citep{forastiere2021identification} defined new causal estimands for treatment and interference in networks and proposed the individual propensity score and neighborhood propensity score by extending the definition of propensity score under neighborhood interference.
The main challenge of estimating causal effects in the network is that the potential outcomes of units in the network depend not only on the treatment assignment but also on the network structure. Units in the network receive interference according to the structures of their treated neighborhoods. It is straightforward to assume units with similar treated neighborhoods will receive similar interference. \\
~\citep{auerbach2021local} proposes a nonparametric modeling framework for causal inference under interference in a sparse network using the configuration of other agents and connections nearby as measured by path distance. A local configuration refers to the features of the network (the agents, their characteristics, treatment statuses, and how they are connected) nearby a focal agent as measured by path distance. This framework assigns distances to subgraphs based on treatment assignments and structural isomorphism. The impact of a policy or treatment assignment is then learned by pooling outcome data across similarly configured agents.
Similarly, numerous methods have been developed to incorporate neighborhood information into estimation procedures, utilizing techniques like graph embedding
Several papers looked into the problem of causal inference under interference in the presence of unobserved confounding and utilized network as a proxy to recover these latent confounders and subsequently adjust for them. ~\citep{veitch_using_2019}  assumes that each person’s treatment and outcome are independent of the network once we know that person’s latent attributes. It only recovers part of the unobserved confounding relevant for the prediction of the propensity score or conditional expected outcome. Then, it plug-in the estimated values of the nuisance parameters to a standard estimator such as A-IPTW estimator to estimate the causal effect. ~\citep{Guo2019LearningIC} extends this by learning representations of hidden confounders through mapping both network structure and features into a shared space, then inferring potential outcomes based on these representations.
~\citep{GIAL} discusses that as network information is incorporated into the model, we face a new imbalance issue,i.e., imbalance of network structure in addition to the imbalance of observed covariate distributions. It is essential to design a new method that can capture the representation of hidden confounders implied from the imbalanced network structure and observed confounders that exist in the covariates simultaneously. To address this issue, the Graph Infomax Adversarial Learning (GIAL) method was introduced. This approach employs Graph Neural Networks combined with structure mutual information to accurately represent both hidden and observed confounders. Following this, a potential outcome generator predicts the potential outcomes for units in both treatment and control groups, using the learned representations and treatment assignments. Concurrently, a counterfactual outcome discriminator is integrated to correct any imbalances between the treatment and control group representations in this learned space.
~\citep{cristali2022using} proposes a method for causal estimation of contagion effects by adjusting for network-inferred attributes without relying on detailed parametric assumptions. They formalize the target causal effect non-parametrically. The main challenge is that the estimand must depend on the network (because contagion requires knowing who is friends with whom) and the network must itself be modeled as a random variable which is a function of the unobserved confounders (to accommodate homophily). Then, they derive sufficient conditions for the estimated attributes to yield causal identification and give a concrete method for contagion estimation using node embedding techniques to extract the information from the network that is relevant for predicting peer influence. This research approach contrasts with our work in this paper, where we operate under the assumption of unconfoundedness. We utilize network information to account for the dependence between units, enabling us to address and adjust for the potentially complex and high-dimensional confounding present within network structures.\\
~\citep{NEM} proposes a non-parametric framework for estimating causal effects under network interference that employs the network embeddings along with matching ~\citep{rosen} to estimate the causal effect. \\
A distinct group of studies focuses on representation balancing to address the issue of differing covariate space distributions among treated and control groups. To avoid biased inference, (~\citep{pmlr-v48-johansson16, pmlr-v70-shalit17a, yao_representation_2018}) propose a balancing counterfactual inference using domain-adapted representation learning. ~\citep{Ma2020CausalIU} extends this by mapping covariate vectors to a feature space, where treated and control groups are balanced through penalizing the distribution discrepancy term (HSIC) between them. This approach is equivalent to finding a feature space such that the treatment assignment and mapped representation become approximately disentangled. Additionally, Graph Neural Networks are employed, followed by an outcome prediction network tailored to the treatment assignment, with a loss function that combines outcome prediction error and distribution
discrepancy in the feature space. In the same category, Similarly, ~\citep{jiang2022estimating} introduces NetEst, which uses GNNs to learn representations of a unit's confounders and those of its neighbors. These representations are then utilized in an adversarial learning process to effectively narrow the distribution gaps between standard graph machine learning and networked causal inference objective function by forcing the mismatched distributions to follow uniform distributions and as a result, accurately estimate the observed outcomes.
~\citep{Guo2020IGNITEAM} combines the ideas of using network as a proxy to learn hidden confounders and balancing the covariate representations across treatment and control group using a minimax game optimization problem. First, a GNN is used to map the features and the adjacency matrix of the network structure into latent space to approximate the confounders, aiming to balance confounder representations across treatment groups to fool the critic. The critic component maps the confounders’ representation of an instance to a real value, with higher values suggesting a higher likelihood of receiving treatment.The objective is to maximize the distinction between treated and controlled instances. Lastly, an outcome inference function tries to infer outcomes of an instance based on its confounders’ representation. \\
The final category in the field of causal effect estimation under interference, and most closely aligned with our research, involves the development of doubly robust estimators as proposed in several studies.
~\citep{mcnealis2023doubly} introduces two novel estimators where the interference set is defined as the set of first-order neighbors assuming that the network is a union of disjoint components. The first estimator, a regression estimator with residual bias correction, is endowed with the double robustness property whether or not the outcome has a multilevel structure. The second estimator, a regression estimator with inverse-propensity weighted coefficients, can be shown to be doubly robust if the outcome does not follow a hierarchical model. This work applies M-estimation theory to propose appropriate asymptotic variance estimators followed by empirical proof of the double robustness and efficiency superiority of these estimators over IPW estimators, even with an accurately specified treatment model. Additionally, the research highlights the risk of latent treatment homophily in identifying causal effects and demonstrates how doubly robust estimation can effectively counter this issue.
~\citep{influencefunc} proposes TMLE, an estimator for treatment and spillover effects and prove asymptotic results under IID assumptions. Finally ~\citep{ogburn2022causal} extends this estimator to allow for dependence due to both contagion and homophily and derive asymptotic results that allow the number of ties per node to increase as the network grows. Their approach utilizes predefined aggregates, an efficient influence function, as introduced by ~\citep{influencefunc}, combined with a moment condition to create a doubly robust estimator. The algorithm employed for estimation resolves the efficient influence function estimating equation through an iterative process.

~\citep{leung2022unconfoundedness} proposes a framework for nonparametric estimation of treatment and spillover effects using observational data from a single large network where interference decays with network distance. They use graph neural networks to estimate the high-dimensional nuisance functions of a doubly robust estimator. They also establish a network analog of approximate sparsity to justify the use of shallow architectures. ~\citep{leung2022unconfoundedness} requires the treatment as well as the neighborhood exposures to be discrete. This can be highly unrealistic in network scenarios. For instance, when studying the effectiveness of vaccination, the protection due to social neighbors' vaccination is often not binary but depends on the proportion of neigbhors vaccinated. Similar is true for social media settings where one is interested in the effectiveness of an ad campaign. Our work allows for the exposures and treatments to be continuous. This is of a prime interest to us especially for our applied example dealing with self-help groups, microinsurance and risk appetite.
Additionally in our research, we calculate two causal estimands: the average direct effect and the average peer effect. This distinction is vital across various real-life scenarios where it's essential to differentiate between the impact of a unit's treatment and that of its neighbors. For example, in the case study highlighted in our paper, we examine the effects of Self-Help Group participation on the financial risk tolerance of both units and their neighboring entities. This analytical separation enhances policy-making insights. In contrast, ~\citep{leung2022unconfoundedness} concentrates solely on the total effect. Furthermore, our theoretical contributions establish guarantees for a semi-parametric estimator using different proof techniques than \citep{leung2022unconfoundedness}.

\section{Causal Inference and
Networks}\label{sec:background}
In this section, we introduce the required notations and the setup of our problem, including causal estimands and necessary assumptions for identification. As a convention in our paper, we represent random variables with capital letters (e.g., $ A$ ), scalars with lowercase letters (e.g., $ a$ ), matrices with script letters (e.g., $ \matr{A}$ ), vectors with boldface symbols (e.g., $ \vect{A}$ ), and sets with blackboard bold symbols (e.g., $ \set{A}$ ). Further, we also denote the shape of the matrix or vector as a subscript when and where necessary e.g. $\matr{A}_{m \times p}$ and $\vect{A}_{m \times 1}$. 
\subsection{Formal Setup and Assumptions}\label{sec:setup}
Consider a social network $ \matr{G} = (\set{V}^n, \matr{A}^n, \matr{Z}^n)$, where $ \set{V}^n = \{1 \dots n\}$ denotes the set of $n$ units, $ \matr{A}^n \in \{0, 1\}^{n \times n}$ is the adjacency matrix representing the connectivity structure of the network across the $n$ units. If $ \matr{A}^n_{i,j} = 1$ for $ i, j \in \set{V}^n$ , then units $i$ and $j$ are connected. The feature matrix $\matr{Z}^n$ can be decomposed as $ \matr{Z}^n = (\matr{X}^n, \vect{T}^n, \vect{Y}^n)$ , where $ \matr{X}^n \in \mathbb{R}^{n \times d}$ is the matrix of pretreatment covariates, $ \vect{T}^n = \{T_1, T_2, \dots, T_n\}$ is the vector of treatments for all units in $\set{V}^n$, and $ \vect{Y}^n = \{Y_1, Y_2, \dots, Y_n\}$ is the vector of outcomes for all units in the network. The potential outcome of unit $i$ under treatment vector $\vect{t}^n$ is denoted by $Y_i(\vect{t}^n)$. We drop the superscript $n$ indicating the sample size for parsimony in the rest of the paper.

Let $ \set{N}_i = \{j : \matr{A}_{i,j} = 1\}$ be the set of nodes sharing ties with node $i$ (i.e., the neighborhood of node $ i$ ). Having `$-i$' in the subscript denotes everything but $i$, hence $ \set{N}_{-i} = \set{V} \setminus (\set{N}_i \cup \{i\})$ is the set of non-neighbors of unit $i$. The vectors of treatments and outcomes for all nodes except node $ i$ are denoted as $ \vect{T_{-i}}$ and $ \vect{Y_{-i}}$ respectively, and the matrix of covariates for all nodes except for node $ i$ as $ \matr{X}_{-i}$. Similarly, for the neighbors of $ i$ , we denote the vectors of their treatments and outcomes as $ \vect{T_{\set{N}_i}}$ and $ \vect{Y_{\set{N}_i}}$ , respectively, and the matrix of their covariates as $ \matr{X}_{\set{N}_i}$.
We assume that our network data is generated via the mechanism defined by the following structural equations:
\setlength{\abovedisplayskip}{10pt}
\setlength{\belowdisplayskip}{10pt}
{
\begin{align} \label{eqn:PLM}  
    T_i &= \phi_T(\vect{X_i}, \matr{X}_{\set{N}_i}) + \epsilon_i^T \quad &\mathbb{E}[\vect{\epsilon^T} \mid \matr{X}] = 0 \nonumber \\
    Y_i &= \theta_0(\vect{X_i}) T_i + \alpha_0(\vect{X_i}) \phi_{YT}(\vect{T_{\set{N}_i}}) + \phi_{YX}(\vect{X_i}, \matr{X}_{\set{N}_i}) + \epsilon_i^Y \quad &\mathbb{E}[\vect{\epsilon^Y} \mid \matr{X}, \vect{T}] = 0   \nonumber
\end{align}
}
\setlength{\abovedisplayskip}{0pt}
\setlength{\belowdisplayskip}{0pt}
\noindent where $ \{\epsilon\}=\{\epsilon^T_i\}_i\cup\{\epsilon^Y_i\}_i$ is a set of unobserved exogenous variables affecting random variables $ \vect{X_i}$, $ T_i$ and $ Y_i$, and $ \phi$'s are a set of functional mappings that describe the causal dependence of the observed variables. 
$\phi_{YX}$ summarizes the covariates of the unit and its peers, i.e. $W_i = \phi_{YX}(\vect{X_i}, \mathcal{X}_{\set{N}_i})$.
Akin to the effective treatment function in ~\cite{manski}, $\phi_{YT}$ is an exposure map that, for any unit $i$, summarizes network peers' treatments $\vect{T}_{\set{N}_i}$ to an effective treatment exposure ~\cite{Aronow2013EstimatingAC}, i.e., $Z_i = \phi_{YT}(\vect{T_{\set{N}_i}})$. In other words, an exposure map is supposed to capture the full nature of interference of a unit from all other units. Given $Z_i$, the outcome $Y_i$ can be determined, rendering it independent of the treatments of the remaining network:  $Y_i(\vect{T}) = Y_i(T_i, Z_i)$.
We operate under the assumption that the exposure mapping $\phi_{YT}$ is well-defined and known. This assumption is common across the literature (see e.g, ~\cite{ogburn2014causal, jiang2022estimating, pmlr-v28-toulis13, zigler2021bipartite, papadogeorgou2023spatial, forastiere2021identification}).

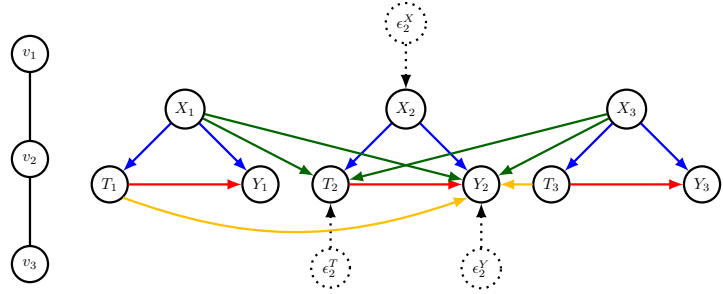
\begin{figure*}[t]
    \centering
    \begin{tikzpicture}[node distance=1.5cm, auto, every node/.style={circle, draw, minimum size=0.8cm}, scale=0.6, transform shape, line width=0.3mm]
    \node (v1) {$ v_1$ };
    \node[below=of v1] (v2) {$ v_2$ };
    \node[below=of v2] (v3) {$ v_3$ };

    \node (X1)[above right=4cm of v3] {$ X_1$ };
    \node[below left=of X1] (T1) {$ T_1$ };
    \node[below right=of X1] (Y1) {$ Y_1$ };

    \node[right=4cm of X1] (X2) {$ X_2$ };
    \node[below left=of X2] (T2) {$ T_2$ };
    \node[below right=of X2] (Y2) {$ Y_2$ };

    \node[below=1cm of T2][dotted] (eT2) {$ \epsilon^T_2$ };
    \node[below=1cm of Y2][dotted] (eY2) {$ \epsilon^Y_2$ };
    \node[above=1cm of X2][dotted] (eX2) {$ \epsilon^X_2$ };

    \node[right=4cm of X2] (X3) {$ X_3$ };
    \node[below left=of X3] (T3) {$ T_3$ };
    \node[below right=of X3] (Y3) {$ Y_3$ };

    \draw[] (v1) -- (v2);
    \draw[] (v2) -- (v3);

    \draw[->, blue, bend left=5] (X1) -- (T1);
    \draw[->, blue, bend right=5] (X1) -- (Y1);
    \draw[->, red, bend right=5] (T1) -- (Y1);

    \draw[->, blue, bend left=5] (X2) -- (T2);
    \draw[->, blue, bend right=5] (X2) -- (Y2);
    \draw[->, red, bend right=5] (T2) -- (Y2);
    \draw[->][dotted] (eX2) -- (X2);
    \draw[->][dotted] (eT2) -- (T2);
    \draw[->][dotted] (eY2) -- (Y2);
    
    \draw[->, blue, bend left=5] (X3) -- (T3);
    \draw[->, blue, bend right=5] (X3) -- (Y3);
    \draw[->, red, bend right=5] (T3) -- (Y3);

    \draw[->,  amber] (T1.south east) to[bend right=20] (Y2.south west);
    \draw[->, amber] (T3.west) to [] (Y2.east);
    \draw[->, green!40!black] (X1) to (T2);
    \draw[->, green!40!black] (X1) to (Y2);
    \draw[->, green!40!black] (X3) to (T2);
    \draw[->, green!40!black] (X3) to (Y2);
\end{tikzpicture}
\caption{Partial causal graph of a network with three nodes. The left side shows the network topology, and the right side depicts the causal graph for each node with $\matr{X}$, $\vect{T}$, and $\vect{Y}$ as confounder, treatment, and outcome, respectively. Solid circles represent endogenous variables; dotted circles, exogenous. Blue edges indicate within-unit confounding, green edges show neighbor confounding, red edges represent direct effects, and yellow edges denote treatment interference.}
    \label{fig:graph}
\end{figure*}

  \textbf{Estimand:} Our objective is to estimate two key causal estimands: the average direct effect (ADE), denoted $ \tau_{\text{ADE}}$, and the average peer effect (APE), denoted $ \tau_{\text{APE}}$. ADE aims to capture the direct impact of treatment on the outcomes within individual units, whereas APE assesses the effect of treatments on a unit through its connections within a network. To illustrate the practical implications of these concepts, consider a friendship network where the treatment is the recommendation of a product in an advertisement to users, and the outcome is the purchasing of the product. This scenario prompts two pertinent questions: How does showing an advertisement to a user influence their likelihood of purchase? And, how does showing an advertisement to a user affect their friends' likelihood of purchase, considering potential discussions about the product? These questions correspond to the ADE and APE, respectively, which are well-established causal estimands in the literature~\cite{hu2022average, jiang2022estimating, halloran1995, RePEc:oup:jeurec:v:19:y:2021:i:4:p:2022-2051., hudgens, sobel}. The estimands are formally defined as follows:
{ \small
\begin{equation}
    \begin{aligned}
    \tau_{\text{ADE}} & = \mathbb{E}_{(\mathcal{X}, \vect{T}, \vect{Y}) \mid  \mathcal{G}} \left[\frac{1}{n} \sum_{i \in \mathbb{V}} \tau_{i,\text{DE}} \right],  \quad \text{where if } t \in \mathbb{R} : \tau_{i,\text{DE}} = \frac{\partial Y_i(t,\vect{T}_{-i})}{\partial t}, \\
    & \quad \quad \text{if } t \in \{0,1\} : \tau_{i,\text{DE}} = Y_i(1,\vect{T}_{-i}) - Y_i(0,\vect{T}_{-i}).
    \end{aligned}
\end{equation}
\begin{equation}
    \begin{aligned}
    \tau_{\text{APE}} & = \mathbb{E}_{(\mathcal{X}, \vect{T}, \vect{Y}) \mid  \mathcal{G}} \left[\frac{1}{n} \sum_{i \in \mathbb{V}} \tau_{i,\text{PE}} \right],  \quad \text{where if } z \in \mathbb{R} : \tau_{i,\text{PE}} = \frac{\partial Y_i(T_i, z)}{\partial z}, \\
    & \quad \quad \text{if } z \in \{0,1\} : \tau_{i,\text{PE}} = Y_i(T_i, \vect{1}_{-i}) - Y_i(T_i, \vect{0}_{-i}).
    \end{aligned}
\end{equation}
}
Here, $ \tau_{i,\text{DE}} $ and $ \tau_{i,\text{PE}} $ respectively denote the direct and peer effects on individual unit $ i $. In the context of the structural equations presented earlier, these correspond to the parameters $ \theta_0(\mathbf{X_i}) $ and $ \alpha_0(\mathbf{X_i}) $. These effects are functions of the pre-treatment variables $ \mathbf{X_i} $. Figure \ref{fig:graph} illustrates a three-node causal graph, demonstrating network dynamics, causal interactions, and the alignment of $ \tau_{i,\text{DE}} $ and $ \tau_{i,\text{PE}} $ within the network structure.

\paragraph{\textbf{Assumptions:}} We introduce the assumptions required for the identification of ADE and APE, which are standard in the causal inference literature from social networks ~\citep{bhattacharya2019causal, guo2020learning, jiang2022estimating, ogburn2022causal}. The network structure, defined by the adjacency matrix $\matr{A}$, is considered fixed and not treated as a random variable. It serves as an information pathway, where connected units can influence each other's treatments and outcomes.
\begin{enumerate}[leftmargin=0.69cm, itemsep=0pt, label = A.\arabic*]
    \item \textbf{Exogeneity:} Unobserved exogenous variables are assumed to be independent. Formally, for any $ i, j \in \mathbb{V}$, we assume:

    \begin{align}
        \vect{\epsilon_i^{\mathcal{X}}} \perp \vect{\epsilon_j^{\mathcal{X}}},
        \epsilon_i^T \perp \epsilon_j^T \mid \vect{X_i}, \vect{X_j}, 
        \epsilon_i^Y \perp \epsilon_j^Y \mid \vect{X_i}, \vect{X_j}, T_i, T_j
    \end{align}

    \item \textbf{Partial Interference:} Each unit's potential outcome is influenced only by its own and its $k$-hop away neighbors' treatments. In this paper, we consider $k = 1$:
    \begin{align}
        Y_i(T_i = t, \vect{T_{\set{N}_i}}, \vect{T_{\set{N}_{-i}}}) = Y_i(T_i = t, \vect{T_{\set{N}_i}}, \vect{T^\prime_{\set{N}_{-i}}})
    \end{align}

    \item \textbf{Known Exposure Map:} The exposure map $\phi_{YT}$ is well-defined and known a priori such that $Z_i = \phi_{YT}(\vect{T_{\set{N}_i}})$
    
    \item \textbf{Positivity:} For all values of $W_i$ present in the population of interest, i.e. $f(W_i) > 0$, all possible values of treatments and exposures have non-zero probabilities:
     
    \begin{align}
        \forall (i, t, z), \; 0 < f(T_i = t, Z_i = z \mid W_i)
    \end{align}
    where $f$ is the probability density function.
    
    \item \textbf{Consistency:} The observed outcomes match potential outcomes under the observed treatment assignments:
    
    \begin{align}
        Y_i(T_i = t, Z_i = z) = Y_i \quad \text{if} \quad T_i = t, Z_i = z.
    \end{align}
    
    \item \textbf{Strong Ignorability:} Conditional on the features $X_i$  and $ \mathcal{X}_{\set{N}_i}$ , the potential outcome is independent of treatment and peer exposure:
     
    \begin{align}
        Y_i(T_i = t_i, Z_i=z) \perp T_i, Z_i \mid X_i,\mathcal{X}_{\set{N}_i}     \end{align}

\end{enumerate}

\begin{proposition} \label{prop: id}
Under the assumptions of A.1-6, the average direct effect (ADE) and the average peer effect (APE) are identifiable.

\end{proposition}
The proof can be found in the appendix \ref{id}.

If the causal effects are constant across units i.e. $\theta_0(\vect{X_i}) = \theta_0$ and $\alpha_0(\vect{X_i}) = \alpha_0$  for all $i \in \set{V}$, then the average direct effect (ADE) is implied by $ \theta_0$, i.e., $ \tau_{\text{ADE}} = \theta_0$, and the average peer effect (APE) is implied by $ \alpha_0$, i.e., $ \tau_{\text{APE}} = \alpha_0$. In this paper, we do not assume heterogeneity in the causal effect; hence, our focus is on estimating $ \theta_0$  and $ \alpha_0$  from network data.

\vspace{-0.3cm}
\section{Method}
In this section, we discuss our estimation strategy for ADE and APE. We operationalize double machine learning machinery with Graph Neural Networks (GNNs) to efficiently estimate ADE and APE by adjusting for complex network confounders. In this section, we illustrate our approach for $\phi_{YT}\left(\vect{T}_{\vect{\set{N}_i}}\right) = \sum_{j \in \vect{\set{N}_i}} T_j$. Recall that,
\begin{equation}\label{eq:exp_1}
        \vect{Y} = \theta_0 \vect{T} + \alpha_0 \mathcal{A} \vect{T} + \phi_{YX}(\vect{X_i}, \matr{X}_{\set{N}_i}) + \vect{\epsilon^Y}.
\end{equation}
Taking expectations with respect to $\matr{X}$ on both sides, noting that $\mathcal{A}$ is constant, and subtracting it from Equation~\ref{eq:exp_1} yields:
    \begin{align} \label{eq:exp}
        \vect{Y}-\ell_0\left(\mathcal{X}, \mathcal{A}\right)= \left(\theta_0 + \alpha_0 \mathcal{A}\right) \cdot\left(\vect{T}-m_0\left(\mathcal{X}, \mathcal{A}\right)\right) + \vect{\epsilon^Y}
    \end{align}
where $m_0\left(\mathcal{X}, \mathcal{A}\right) := \mathbb{E}\left[\vect{T} \mid \mathcal{X}, \mathcal{A}\right]$ and $\ell_0\left(\mathcal{X}, \mathcal{A}\right) := \mathbb{E}\left[\vect{Y} \mid \mathcal{X}, \mathcal{A}\right]$. 

Let $\zeta = \left(\theta, \alpha\right)$ and $\eta = \left(m, \ell\right)$ to be the unknown target and nuisance parameters with $\zeta_0 := \left(\theta_0,\alpha_0\right)$ and $\eta_0 := \left(m_0,\ell_0\right)$ as the true values of these parameters of our interest that satisfies equation \ref{eq:exp}. Now, let $W := \left(\matr{X},\vect{T},\vect{Y}\right)$ be a random element taking values in a measurable space $\left(\mathcal{W},\mathcal{A}_\mathcal{W}\right)$ with law determined by a probability measure $P \in \mathcal{P}_N$ with $\left(W_i\right)_{i=1}^n$ random samples available for estimation and inference. Then, consider a squared loss derived
        $\mathcal{L}\left(W, \matr{A} ; \zeta, \eta\right) := \frac{\vect{B}^\intercal_{1 \times n} \vect{B}_{n \times 1}}{2}$
 where $\vect{B}_{n \times 1} := \left[\vect{Y} - \vect{\ell\left(\mathcal{X}, \mathcal{A}\right)} - \left(\theta + \alpha \mathcal{A}\right)\left(\vect{T} - \vect{m\left(\mathcal{X}, \mathcal{A}\right)}\right) \right]$ such that the partial derivatives of the loss function with respect to target parameters and nuisance parameters, evaluated at $\zeta_0$ and $\eta_0$ yields zero:
\begin{align*}
\mathbb{E}_P\left[\left.\partial_\zeta \mathcal{L}\left(W, \matr{A} ; \zeta, \eta\right)\right\vert_{\zeta_0, \eta_0}\right]=0 , \quad 
\mathbb{E}_P\left[\left.\partial_\eta \mathcal{L}\left(W, \matr{A} ; \zeta, \eta\right) \right\vert_{\zeta_0, \eta_0}\ \right]=0
\end{align*}
Thus, the target parameters can be identified by minimizing the following squared loss:
\begin{equation*}
\zeta_0, \eta_0 \in \arg \min_{\zeta, \eta} \mathbb{E}_P \left[\mathcal{L}\left(W, \matr{A} ; \zeta, \eta\right)\right],
\end{equation*}
Now, we construct an efficient score function, $\psi$ that enables doubly robust estimation of target parameters, similar to \cite{chernozhukov2018double} and \cite{morucci2023double}:
\begin{equation*}
\psi\left(W, \matr{A} ; \zeta, \eta\right) = \partial_\zeta \mathcal{L}\left(W, \matr{A} ; \zeta, \eta\right)-\mu \partial_\eta \mathcal{L}\left(W, \matr{A} ; \zeta, \eta\right),
\end{equation*}
where $\mu$ is an orthogonalization parameter matrix such that its optimal value solves the equation: $
J_{\zeta \eta}-\mu J_{\eta \eta}=0
$ where,
\begin{equation*}
    \left(\begin{array}{cc}
    J_{\zeta \zeta} & J_{\zeta \eta} \\
    J_{\eta \zeta} & J_{\eta \eta}
    \end{array}\right)=\left.\partial_{\left(\zeta^{\prime}, \eta^{\prime}\right)} \mathbb{E}_P\left[\partial_{\left(\zeta^{\prime}, \eta^{\prime}\right)^{\prime}} \mathcal{L}\left(W ; \zeta, \eta\right)\right]\right|_{\zeta_0 ; \eta_0}.
\end{equation*}
The detailed derivation of the score function is provided in the appendix~\ref{score} for further reference. The score function is identified as follows:
\begin{equation}
    \begin{aligned}
        \begin{split}
        \psi\left(W, \matr{A} ; \zeta, \eta\right) &= 
        \begin{pmatrix}
        (\vect{Y} - \vect{\ell\left(\mathcal{X}, \mathcal{A}\right)} - \left(\theta + \alpha \mathcal{A}\right)\left(\vect{T} - \vect{m\left(\mathcal{X}, \mathcal{A}\right)}\right))^\intercal\left(\vect{T} - \vect{m\left(\matr{X}, \mathcal{A}\right)}\right) \\
        (\vect{Y} - \vect{\ell\left(\mathcal{X}, \mathcal{A}\right)} - \left(\theta + \alpha \mathcal{A}\right)\left(\vect{T} - \vect{m\left(\mathcal{X}, \mathcal{A}\right)}\right))^\intercal \mathcal{A} \left(\vect{T} - \vect{m\left(\matr{X}, \mathcal{A}\right)}\right)
        \end{pmatrix}
        \end{split}.
    \end{aligned}
\end{equation}

We can now use the score function to construct an estimator for $\zeta$ such that $\psi\left(W, \matr{A}; \widehat{\zeta},\widehat{\eta}\right) = 0$ where $\widehat{\eta} = \left(\widehat{\ell},\widehat{m}\right)$ are the estimates of nuisance parameters. Thus, 
\begin{eqnarray*}
    && \left[\left(\vect{Y} - \vect{\widehat{\ell}\left(\mathcal{X}, \mathcal{A}\right)}\right)^\intercal \left(\vect{T} - \vect{\widehat{m}\left(\mathcal{X}, \mathcal{A}\right)}\right)\right] = \widehat{\theta} \left[\left(\vect{T} - \vect{\widehat{m}\left(\mathcal{X}, \mathcal{A}\right)}\right)^\intercal \left(\vect{T} - \vect{\widehat{m}\left(\mathcal{X}, \mathcal{A}\right)}\right)\right] + \\&&\quad \widehat{\alpha} \left[\left(\vect{T} - \vect{\widehat{m}\left(\mathcal{X}, \mathcal{A}\right)}\right)^\intercal \mathcal{A}^\intercal \left(\vect{T} - \vect{\widehat{m}\left(\mathcal{X}, \mathcal{A}\right)}\right)\right]
\end{eqnarray*}
and
\begin{eqnarray*}
    && \left[\left(\vect{Y} - \vect{\widehat{\ell}\left(\mathcal{X}, \mathcal{A}\right)}\right)^\intercal \mathcal{A}\left(\vect{T} - \vect{\widehat{m}\left(\mathcal{X}, \mathcal{A}\right)}\right)\right] = \widehat{\theta} \left[\left(\vect{T} - \vect{\widehat{m}\left(\mathcal{X}, \mathcal{A}\right)}\right)^\intercal \mathcal{A}\left(\vect{T} - \vect{\widehat{m}\left(\mathcal{X}, \mathcal{A}\right)}\right)\right] + \\&&\quad  \widehat{\alpha} \left[\left(\vect{T} - \vect{\widehat{m}\left(\mathcal{X}, \mathcal{A}\right)}\right)^\intercal\mathcal{A}^\intercal\mathcal{A} \left(\vect{T} - \vect{\widehat{m}\left(\mathcal{X}, \mathcal{A}\right)}\right)\right].
\end{eqnarray*}

For accurate and consistent estimation of nuisance parameters $\eta_0$, we leverage graph machine learning approaches, specifically using GNNs. Since nuisance parameters $m$ and $\ell$ depend on both individual unit covariates and their social neighbors' covariates, GNNs are essential for aggregating neighborhood information. GNNs efficiently handle the complex dependencies in network data, as shown in \cite{xu2018powerful, kipf2016semi, velivckovic2017graph, hamilton2017inductive}. In our experiments, we use the Graph Isomorphism Network (GIN) \cite{xu2018powerful} due to its superior performance over other GNN architectures like GCN \cite{kipf2016semi}, GAT \cite{velivckovic2017graph}, and GraphSAGE \cite{hamilton2017inductive}.  
For consistent estimation of nuisance parameters and to address non-i.i.d. data, we use a focal set approach similar to~\cite{RePEc:nbr:nberwo:21313}. Our algorithm first constructs a set of units that are independent of each other, referred to as the focal set, defined formally below, and then performs cross-fitting to train the graph machine learning model for modeling the nuisance parameters on the focal set. This independence across units aids in consistent estimation of uncertainty around the estimated target parameters by avoiding bias due to dependence between the units. 

Below, we formally define the focal set:
\begin{definition}
\label{def:focal}
Focal set $\mathbb{S}^* \subseteq \mathbb{V}$ is a maximal set of nodes, in which  $\forall u, v \in \mathbb{S}^*$, $\set{N}_u \cap \set{N}_v = \emptyset$. We denote the size of the focal set  $\mathbb{S}^*$ as $n_f$, i.e. $|\mathbb{S}^*| = n_f$
\end{definition}
According to the partial interference assumption, since neighborhoods of nodes in the focal set do not overlap, $\left(\mathcal{X},\vect{T},\vect{Y}\right)$ of these nodes will not be correlated, ensuring that the samples are independent of each other.

Further, as discussed by~\citep{chernozhukov2018double, zivich2021machine, parikh2022malts}, for the error term of the estimator to vanish, to overcome overfitting, and to gain full efficiency, we cross-fit our estimator.
Consider a $K-$fold random partition $\left(I_k\right)_{k=1}^K$ of our data $\{1,...,n_f\}$, such that each fold $I_k$ will be of size $\frac{n_f}{K}$. Let $I_{-k} = \{1,...,n_f\} \backslash I_k$. For each $k$, let $I_{-k}$ be the train split and $I_{k}$ be the estimation split.  We construct a ML estimator $\hat{\eta}_{k}$ of the nuisance function $\eta_0$ using the train split:
\begin{equation}
\hat{\eta}_{k} = \hat{\eta}\left(\left(W_i\right) _{i \in I_{-k}}\right).
\end{equation}
Then, for each $k \in \{1\dots K\}$, we plugin the estimated nuisance parameters $\hat{\eta}_{k}$  to estimate $\hat{\zeta}_{k}$ as the solution to 
$\frac{K}{n_f} \sum_{i \in I_k} \psi\left(W_i, \matr{A}; \zeta, \hat{\eta}_{k}\right) = 0$

Our final estimation would be an aggregation of the estimators:
$\widehat{\zeta} = \frac{1}{K}\sum_{k=1}^K \hat{\zeta}_{k}$.
Note that the choice of $K$ may have a significant impact in small sample sizes. Intuitively, selecting larger values of $K$ yields more observations in $I_{-k}$, which can be advantageous for estimating high-dimensional nuisance functions, which seems to be the more difficult part of the problem. Empirical evidence and simulations indicate that moderate values of $K$, such as 4 or 5, yield more reliable estimations than using $K = 2$. This underscores the importance of carefully selecting $K$ based on the sample size and the complexity of the functions being estimated.

\begin{figure*}[t]
    \centering
    \includegraphics[width=1.0\textwidth]{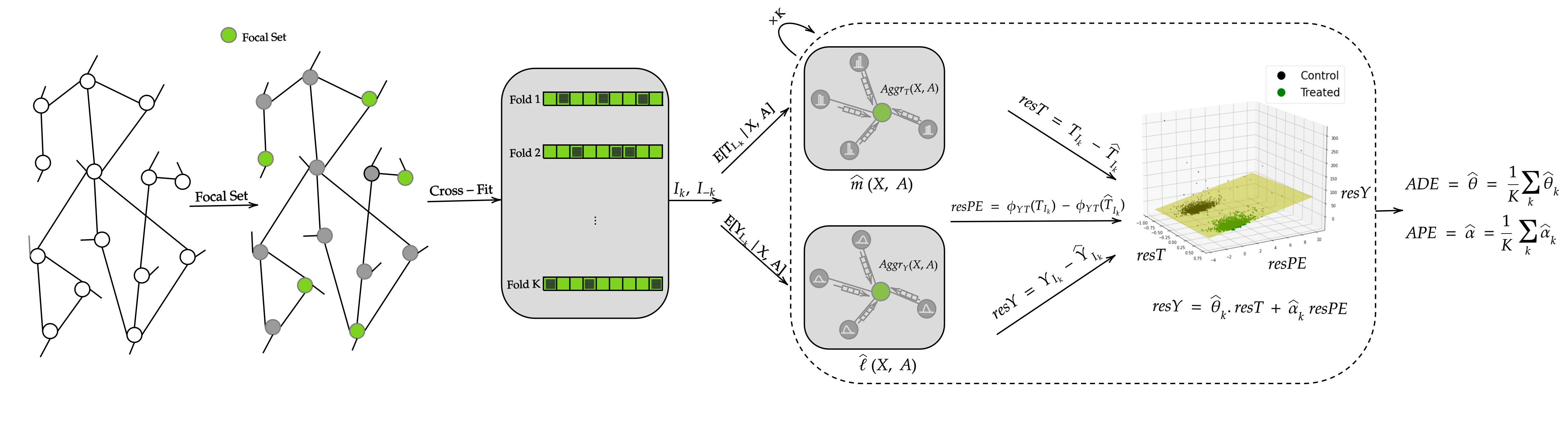}
    \caption{ Framework schema. The focal set is partitioned into train and estimation folds $I_{-k}$ and $I_k$ for cross-fitting. Propensity score and outcome models are learned over $I_{-k}$ using graph machine learning. Estimations of $\mathbb{E}[T \mid X, A]$ and $\mathbb{E}[Y \mid X, A]$ for $I_k$ are computed to derive residuals $resT$, $resPE$, and $resY$. Finally, $resY$ is regressed on $resT$ and $resPE$ to obtain $\hat{\theta_k}$ and $\hat{\alpha_k}$ for ADE and APE. This process is repeated across folds, and results are aggregated for final estimations of $\theta$ and $\alpha$.
    }
    \label{fig:diagram}
\end{figure*}

\textbf{Putting Everything Together.}
To estimate ADE and APE, our method begins by constructing a 'focal set' of units using a greedy approach to create an independent set of nodes. This focal set is the core of our analysis. Using graph machine learning, we train models on this focal set to accurately model the nuisance functions. To enhance accuracy and robustness, we employ cross-fitting by partitioning the data into multiple folds. In each fold, we perform linear regression to estimate the parameters $\theta_0$ and $\alpha_0$, representing the direct and peer effects, respectively. These estimations are then aggregated across all folds to construct our comprehensive model of network dynamics. This integrative approach, combining the precision of graph machine leaning with the robustness of cross-fitting and the targeted analysis of the focal set, enables a nuanced and precise understanding of causal relationships in social networks. Figure \ref{fig:diagram} illustrates the proposed framework.

\section{Theory}
Now, we establish the theoretical results on the consistency and asymptotic normality of the proposed estimator. Detailed proofs are provided in Appendix~\ref{sec: dml_cond}. We consider a nested sequence of networks with an increasing number of units, $\{ \set{V}^n, \matr{A}^n, \matr{Z}^n \}_{n=1}^{\infty}$, such that key features of the network topology, e.g. degree distribution and clustering,
are preserved. We assume that the maximum degree of the units in $\matr{A}^n$ is $d_n \leq \sqrt{n}-1$. This growth rate of the maximum degree of the network is a common trait in many real-world social networks where most units possess a low degree, and a smaller proportion of units have a high degree, with the maximum degree dependent on the size of the network~\citep{newman2003social}. This characteristic ensures broad applicability of our model to real-world social networks.

To prove the theoretical results, we need some regularity conditions adopted from the DML framework \citep{chernozhukov2018double} and adapted to social networks. These conditions capture sufficient assumptions about the behavior and properties of the data and models to ensure stable and valid inference. Intuitively, these conditions ensure enough variability in the treatment and outcome models, prevent the alignment of error terms that could distort causal effect estimation, and ensure accurate and reliable estimators for nuisance parameters. Additionally, they guarantee that the nuisance parameter estimators converge to their true values as the sample size increases, which is crucial for the consistency and asymptotic normality of the causal estimators, allowing for valid statistical inference. 

Now, we discuss the theoretical results supporting the consistency and asymptotic normality of the proposed estimator. To maintain a clear and focused narrative in the main text, we have relegated all the detailed proofs to the appendix~\ref{sec: dml_cond}.
We consider a nested sequence of networks with an increasing number of units, $\{ \set{V}^n, \matr{A}^n, \matr{Z}^n \}_{n=1}^{\infty}$, such that key features of the network topology, e.g. degree distribution and clustering,
are preserved. We assume that the maximum degree of the units in $\matr{A}^n$ is $d_n \leq \sqrt{n}-1$. This growth rate of the maximum degree of the network is a common trait in many real-world social networks where most units possess a low degree, and a smaller proportion of units have a high degree, with the maximum degree dependent on the size of the network \citep{newman2003social}. This characteristic ensures that our model remains applicable to a wide range of real-world social networks. 

Now, we assume the following regularity conditions: 
\begin{assumption} \label{a: regconds}
    (Regularity Conditions) Let $c > 0, C > 0, c_1 \geqslant c_0 > 0, q > 4$ and $K \geqslant 2$ be some finite constants, and let $\{\delta_n\}_{n = 1}^{\infty}$ and $\{\Delta_n\}_{n = 1}^{\infty}$ be some sequences of positive constants converging to zero such that $\delta_n \geqslant n_f^{-1/2}$. For all probability laws $P \in \mathcal{P}$ for the triple $W = (\vect{T}, \vect{Y}, \mathcal{X})$ the following
    conditions hold:
    \begin{enumerate}
        \item $c \leqslant \|\vect{\epsilon^T}\|_{P,2}$ ; $ \|\vect{\epsilon^T}\|_{P,q} \leqslant C$; $\leqslant \|\vect{\epsilon^Y}\|_{P,q} \leqslant C$
        \item $c \leqslant {\| \vect{\epsilon^Y}^\intercal \vect{\epsilon^T} \|}_{P,2}$; 
        $c \leqslant \mathbb{E}_P\left[\vect{\epsilon^T}^\intercal \vect{\epsilon^T} \right]$; \\
        $c \leqslant \mathbb{E}_P\left[\vect{\epsilon^T}^\intercal \mathcal{A}^\intercal \vect{\epsilon^T} \right]$
        \item $\|\vect{Y}\|_{P,q} \leqslant C$
        \item $\vect{\epsilon^T}$ and $ \vect{\epsilon^Y}$ are not eigen vectors of $\mathcal{A}$.
        \item Given a random subset $I$ of $\{1\dots,n_f\}$ of size $n^{\prime}=n_f / K$, the nuisance parameter estimator $\widehat{\eta}=\widehat{\eta}\left(\left(W_i\right)_{i \in I_{-k}}\right)$ belongs to the realization set $\Gamma_{n}$ with probability at least $1-\Delta_{n}$, where $\eta_0 \in \Gamma_{n}$.
        \item With $P$-probability no less than $1-\Delta_N$,
$$
\left\|\widehat{\eta}-\eta_0\right\|_{P, q} \leqslant C, \quad\left\|\widehat{\eta}-\eta_0\right\|_{P, 2} \leqslant \delta_n, \quad \text { and }
$$
for the score $\psi$, where $\widehat{\eta}_0=\left(\widehat{m}_0, \widehat{\ell}_0\right)$,
\small{$$
\left\|\widehat{m}-m_0\right\|_{P, 2} \left(\left\|\widehat{m}-m_0\right\|_{P, 2}+\left\|\widehat{\ell}-\ell_0\right\|_{P, 2}\right) \leqslant \frac{\delta_n}{n_f^{1 / 2}} .
$$}
\end{enumerate}
\end{assumption}

\begin{theorem}\label{th: theory}
   Under regularity conditions \ref{a: regconds}\footnote{These conditions are discussed in more depth in Appendix~\ref{sec: dml_cond}}, the estimator $\vect{\tilde{\zeta}_0}$ concentrates in a $\sigma / \sqrt{n_f}$-neighborhood of $\vect{\zeta_0}$ and the sampling error $\sqrt{n_f}\left(\vect{\tilde{\zeta}_0}-\vect{\zeta_0}\right)$ is asymptotically normal
$$
\sqrt{n_f}\left(\vect{\tilde{\zeta}_0}-\vect{\zeta_0} \right) \rightsquigarrow N\left(\vect{0}_{2 \times 1}, \sigma_{2 \times 2}^2\right)
$$
with mean zero and variance given by
$$
\sigma^2:=(\vect{J_0})^{-1} \mathbb{E}\left[\psi\left(W ; \vect{\zeta_0}, \vect{\eta_0}\right) \psi\left(W ; \vect{\zeta_0}, \vect{\eta_0} \right)^{\intercal}\right] ((\vect{J_0})^{-1})^{\intercal}
$$
where $\vect{J_0}=\mathbb{E}\left(\psi_a\left(W ; \vect{\eta_0} \right)\right)$, if the score function is linear in the parameters $\vect{\zeta}$. For these score functions, estimates of the variance, $\hat{\sigma}^2$, are obtained by
{\small
\begin{eqnarray*}
&& (\vect{\hat{J}_0})^{-1} \frac{1}{n_f}\sum_{k=1}^K \sum_{i \in I_k} \biggl[\psi\left(W_i ; \vect{\tilde{\zeta}}, \vect{\hat{\eta}}_{k}\right)
 \psi\left(W_i ; \vect{\tilde{\zeta}}, \vect{\hat{\eta}}_{k}\right)^{\intercal}\biggl] ((\vect{\hat{J}_0})^{-1})^{\intercal},\\&& \text{ where }
\vect{\hat{J}_0} =\frac{1}{n_f} \sum_{k=1}^K \sum_{i \in I_k} \psi_a\left(W_i ; \vect{\hat{\eta}}_{k}\right)
\\
&&\psi_a = \left(\begin{array}{cc}
-(\vect{T} - \vect{\widehat{m}(\matr{X}, \mathcal{A})})^\intercal(\vect{T} - \vect{\widehat{m}(\matr{X}, \mathcal{A})}) & 
-(\vect{T} - \vect{\widehat{m}(\matr{X}, \mathcal{A})})^\intercal \mathcal{A}^\intercal(\vect{T} - \vect{\widehat{m}(\matr{X}, \mathcal{A})}) \\
-(\vect{T} - \vect{\widehat{m}(\matr{X}, \mathcal{A})})^\intercal \mathcal{A}(\vect{T} - \vect{\widehat{m}(\matr{X}, \mathcal{A})}) & 
-(\vect{T} - \vect{\widehat{m}(\matr{X}, \mathcal{A})})^\intercal \mathcal{A}^\intercal \mathcal{A}(\vect{T} - \vect{\widehat{m}(\matr{X}, \mathcal{A})})
\end{array}\right)\end{eqnarray*}}
The confidence interval is given by
$$
\left[\vect{\tilde{\zeta}_0} \pm \hat{\sigma} / \sqrt{n_f} Z^{-1}(1-\alpha / 2)\right]
$$
 
\end{theorem}

The result of Theorem~\ref{th: theory} guarantees that our estimator is consistent, asymptotically normal, and statistically efficient in the size of the focal set such that the standard deviation shrinks at the rate of $\sqrt{n_f}$. The proof of the theorem is in Appendix~\ref{sec: dml_cond}

\section{Empirical Analysis and Results} \label{sec:results}

This section details the empirical evaluation of our framework via semi-synthetic and real-data case studies. These experiments aim to examine our framework's effectiveness and compare its performance with state-of-the-art baseline methods. In Appendix~\ref{sec: additional_exps}, we present additional empirical experiments to investigate the performance of our approach under varying levels of graph density and corresponding effective sample sizes. We evaluate the coverage probability of estimated 95\% confidence intervals and explore the performance of an alternative graph aggregation tool combined with our framework, demonstrating the framework's generality beyond GNN models.

\begin{table}[t]
    \centering
    
    \resizebox{\columnwidth}{!}{
    \begin{tabular}{@{}lccccccccc@{}}
        \toprule
        & \multicolumn{3}{c}{Cora} & \multicolumn{3}{c}{Pubmed} &\multicolumn{3}{c}{Flickr} \\
        \cmidrule(lr){2-4} \cmidrule(lr){5-7} \cmidrule(lr){8-10}
        & \multicolumn{1}{c}{ADE} & \multicolumn{1}{c}{APE} & \multicolumn{1}{c}{ATE} & \multicolumn{1}{c}{ADE} & \multicolumn{1}{c}{APE} & \multicolumn{1}{c}{ATE} & \multicolumn{1}{c}{ADE} & \multicolumn{1}{c}{APE} & \multicolumn{1}{c}{ATE}\\
        \midrule
        
        PA         & ${0.31}_{\scriptsize{\pm {0.83}}}$
 & $1.02_{\scriptsize{\pm 2.90}}$ & $1.41_{\scriptsize{\pm 4.05}}$ & $0.35_{\scriptsize{\pm 0.69}}$ & $10.69_{\scriptsize{\pm 6.52}}$ & $14.30_{\scriptsize{\pm 9.66}}$ & $1133_{\scriptsize{\pm 4700}}$ & $37719_{\scriptsize{\pm 143300}}$ & $51790_{\scriptsize{\pm 199836}}$\\
        T-learner\citep{tlearner} & $9.84_{\scriptsize{\pm 51.32}}$ & N/A & N/A & $1.67_{\scriptsize{\pm 4.58}}$ & N/A & N/A &$2380_{\scriptsize{\pm 5495}}$& N/A & N/A\\
        NetEst\citep{jiang2022estimating}  & $174.66_{\scriptsize{\pm 1.07}}$
 & $9.48_{\scriptsize{\pm 1.75}}$ & $71.96_{\scriptsize{\pm 4.35}}$ & $1655.8_{\scriptsize{\pm 30.94}}$ & $0.44_{\scriptsize{\pm 0.39}}$ & $1603.53_{\scriptsize{\pm 45.65}}$ & $53827_{\scriptsize{\pm 921}}$ & ${103}_{\scriptsize{\pm {105}}}$ & $58503_{\scriptsize{\pm 3444}}$\\
        Net TMLE\citep{ogburn2022causal} & $13.67_{\scriptsize{\pm 6.47}}$
 & N/A & N/A & $1.24_{\scriptsize{\pm 1.36}}$ & N/A & N/A & N/A* & N/A & N/A\\
        L\&L\citep{leung2022unconfoundedness} & N/A & N/A & $120.20_{\scriptsize{\pm 34.31}}$ & N/A & N/A & $42.76_{\scriptsize{\pm 2.92}}$ & N/A & N/A & ${25.32}_{\scriptsize{\pm {6.58}}}$ \\
        Ma \& Tresp\citep{Ma2020CausalIU} & $3.87_{\scriptsize{\pm 42.98}}$ & ${0.02}_{\scriptsize{\pm {0.14}}}$  & $4.26_{\scriptsize{\pm 47.37}}$ & ${0.02}_{\scriptsize{\pm {0.03}}}$ & ${0.01}_{\scriptsize{\pm {0.01}}}$ & ${0.04}_{\scriptsize{\pm {0.06}}}$ & N/A* & N/A* & N/A*\\
        \midrule
        GDML w/o FS & ${0.26}_{\scriptsize{\pm {0.83}}}$ & $0.99_{\scriptsize{\pm 2.79}}$  & ${1.37}_{\scriptsize{\pm {3.72}}}$ &$0.04_{\scriptsize{\pm 0.13}}$ & $0.30_{\scriptsize{\pm 0.73}}$ & $0.45_{\scriptsize{\pm 1.21}}$ & ${4.92}_{\scriptsize{\pm {10.91}}}$
 & $121_{\scriptsize{\pm 308}}$
 & $95_{\scriptsize{\pm 276}}$\\
        GDML  & $0.33_{\scriptsize{{\pm 0.79}}}$
 & ${0.29}_{\scriptsize{\pm {0.80}}}$
 & ${0.88}_{\scriptsize{\pm {2.21}}}$ & ${0.03}{\scriptsize{\pm {0.11}}}$ & ${0.28}_{\scriptsize{\pm {0.84}}}$ & ${0.30}_{\scriptsize{\pm {0.87}}}$
 & ${76}_{\scriptsize{\pm {211}}}$
 &  ${26.01}_{\scriptsize{\pm {26.07}}}$ & ${84}_{\scriptsize{\pm {272}}}$
\\
        \bottomrule
    \end{tabular}}
    \vskip 0.1in
       \caption{Comparison of mean squared error of our GDML approach with other baselines.
       For T-Learner and Net TMLE methods, peer effect estimation is not applicable. L\&L's framework concentrates on total effect and does not calculate ADE and APE separately.\\
 *: results for Net TMLE and MaTresp on Flickr are not reported because it ran out of system memory.}
  \label{table:baselines}
\end{table}



\begin{table}[ht]
    \centering
    \resizebox{\columnwidth}{!}{
    \begin{tabular}{@{}lcccccccccccccc@{}}
        \toprule
        \multirow{2}{*}{Dataset} & \multicolumn{8}{c}{Method} \\
        \cmidrule(l){2-9} 
        & PA         & T-learner\citep{tlearner} & NetEst\citep{jiang2022estimating} & Net TMLE\citep{ogburn2022causal} & L\&L\citep{leung2022unconfoundedness}  & Ma \& Tresp\citep{Ma2020CausalIU} & GDML w/o FS & GDML \\
        \midrule
        Cora  & 2  & 3.32  & 19020  & 4  & 5  & 26 & 9  & 5 \\
        Pubmed  & 55  & 48  & 22560  & 2464 & 133  & 8604 & 104  & 64 \\
        Flickr  & 1666 & 1712  & 31118  & N/A  & 2909  & N/A  & 2296  & 1832 \\
        \bottomrule
    \end{tabular}}
    \vskip 0.1in
    \caption{Comparison of runtime in seconds for different methods on each dataset. N/A indicates that the method did not return any results within a 12-hour runtime limit.}
    \label{table:time}
    \begin{flushleft}
    \end{flushleft}
\end{table}

\vspace{-0.3cm}
\paragraph{Setup:} \label{exp: setup}
We use real-world networks from the Cora \citep{mccallum2000automating}, Pubmed \citep{sen2008collective}, and Flickr \citep{guo2020learning} datasets. To access potential outcomes, we use the networks from these datasets and generate semi-synthetic data with synthetic covariates, treatments and outcomes, ensuring ground truth availability. Details of these datasets and the data generation processes are provided in Appendix~\ref{dataset_appendix} and~\ref{dgp_appendix}. While our method can be used with any graph learning algorithm, we employ the GIN~\citep{douglas2011weisfeiler} for learning the propensity score and outcome models, which outperforms other methods we tested. We use a single layer of $\operatorname{GINConv}$, as per the partial interference assumption, followed by two fully connected layers and an additional softmax layer for estimating the propensity score. Nonlinearity is introduced using $\operatorname{ReLU}$, and dropout with a probability of $p=0.5$ is used for regularization. The GIN models are trained over 300 epochs with a batch size of 16, using the Adam optimizer with a learning rate of 0.01. We consider $K=3$ folds for cross-fitting. The codebase will be publicly available for further exploration and reference. Experiments on Cora and Pubmed are executed on MacBook Pro 18 with Apple M1 Pro chip, 10 CPUs (8 performance and 2 efficiency) and 16 GB RAM. The experiments on Flickr are executed on the server with GPU model NVIDIA RTX-3090.

\paragraph{Baselines:} We compare our method against six primary baselines. NetEst \citep{jiang2022estimating} uses GNNs for learning confounder representations with an adversarial learning process. Net-TMLE \citep{ogburn2022causal} derives a doubly robust estimator using an efficient influence function and moment condition. The T-Learner \citep{tlearner} creates two estimators for each treatment arm using GNNs for modeling. DML with predefined aggregates applies DML in the i.i.d. setting using aggregates like max, min, and mean for neighbor information. Ma \& Tresp method~\citep{Ma2020CausalIU} maps covariate representations to a new space using HSIC as a regularization term, with GNNs aggregating neighbor covariate information. Lastly, the L\&L method \citep{leung2022unconfoundedness} utilizes a doubly robust estimator combined with GNNs, requiring binary conversion of exposure data (this is a working paper with no public available code, hence we implemented it ourselves; details are in the Appendix~\ref{ex:baselines}). Similarly, \citep{guo2020learning} captures hidden confounder influence but reduces to the T-learner under our assumptions.

\paragraph{Results:} We compare two versions of our method: GDML w/o FS, which does not use a focal set and encompasses the entire dataset, and GDML, which operationalize our method using focal set to evaluate the effect of this strategy on the quality of the results. Table~\ref{table:baselines} demonstrates the results, and Table~\ref{table:time} reports the running time. We report the mean squared error (MSE) calculated over 100 simulations for each evaluated method (relative errors are reported in Figure \ref{fig:baselines} in the appendix).

Across all three semi-synthetic datasets, our GDML approach performs on par with or better than the state-of-the-art baseline methods (see Table~\ref{table:baselines}) and scales significantly better (see Table~\ref{table:time}). This performance enhancement can be attributed to the use of graph ML method (such as GIN) to adjust for network confounders. Further, the DML framework guarantees consistency, and efficiency while using complex ML methods with regularization. While Tresp \& Ma and L\&L occasionally have MSE marginally smaller, our approach consistently delivers robust results across various scenarios while being scalable and running in a reasonable time. Comparing GDML with the GIN-based T-learner shows that the GIN-based T-learner has higher MSE due to regularization-induced bias. Methods that employ predefined aggregation functions, such as Net-TMLE and DML methods using predefined aggregates (min, max, sum, and average), fall short as these aggregates do not capture complex network functions as effectively as GNNs. Tresp \& Ma, while slightly better in certain cases, fails to scale well, taking more than two and a half hours on PubMed and failing to terminate on Flickr within a 12-hour time limit. Additionally, Tresp \& Ma does not provide confidence intervals, which prevents us from performing uncertainty quantification—a key aspect when applying real-world datasets without access to the true data generative process. Among all these eight methods, only our GDML approach, Net-TMLE, and L\&L also yield consistent confidence intervals necessary for statistical inference.

A pivotal aspect of our methodology is the emphasis on focal sets analysis, exploiting the independence between units to enhance performance. While restricting to local sets reduces the size of training data for nuisance parameter estimation, this focus has demonstrably outperformed variants of our method that omit focal sets, highlighting the strategic value of considering focal sets in the analysis of networked data. This ensures robust and accurate estimations, validating our approach in dealing with networked data.
\vspace{-0.3cm}
\paragraph{Analysis of Real Data} \label{exp: casestudy}
For a case study, we used the Indian Village dataset from Karnataka, India, encompassing 16,995 individuals across 77 villages, with 15 features and 12 social networks \citep{banerjee2014gossip, jackson2012social}. This dataset's rich social structure provides insights into economic and social behaviors, such as borrowing, lending, and advice networks, making it invaluable for understanding the impacts of social networks on individual and collective outcomes. We operationalize our approach to estimating the causal effect of participation in self-help groups (SHGs) on financial risk tolerance, measured by the existence of an outstanding credit/loan. Specifically, we are interested in estimating the direct and peer effects of SHG participation. We construct the focal set that consists of 1766 individuals.
Our analysis indicates that the point estimate for the average direct effect (ADE) is $0.315$ with a $95\%$ confidence interval ranging from $-1.570$ to $2.200$. The positive point estimate suggests a potential positive effect of SHG participation. 
However, given the limited effective sample size in the social network setting (which is equal to the size of the focal set in our case), these estimates are not statistically significant, which is expected as the potential effect size in such social interventions is typically small. 
Additionally, our results show that the APE is approximately zero ($0.050$), indicating minimal to no benefit from peers' participation in SHGs. We provide point estimates from the baselines in Appendix\ref{casestudy} -- our estimates agree with these point estimates as well as with the literature.

\section{Conclusions and Broader Impacts} \label{sec: conclusion}
Our work proposes a double-machine learning framework integrated with graph representation learning techniques to adjust for complex network confounders and efficiently estimate treatment effects. Evaluations through thorough simulation studies and real data case studies highlight its effectiveness. However, our framework has limitations, such as reliance on assumptions about GNN convergence rates, which are still an emerging area of research. Additionally, it requires the observation of all confounders, albeit accommodating high-dimensional covariates with complex mappings for propensity scores and outcome models, and necessitates domain-specific knowledge about exposure mappings. The efficacy of our framework diminishes with densely connected graphs due to the reduction in effective sample size. Future directions include adapting the framework for relational data scenarios with heterogeneous graphs, exploring higher-order graphs to enhance expressiveness beyond the limitations of message-passing GNNs, and investigating the impact of missing network ties on estimation in partially observable graphs, aiming to broaden the scope and applicability of our framework.

Our research improves causal inference in social networks, aiding policy-making with more accurate evaluations of intervention effectiveness. This enhances public health, education, and economic policies. It also supports economic development by identifying effective social programs like self-help groups, which promote financial inclusion and risk management in underserved communities. Nonetheless, ethical, transparent, and correct application of these tools is crucial to avoid privacy issues and potential biases. \label{socimp}

\bibliographystyle{apalike}  
\bibliography{references}

\section{Proof of Identifiability} \label{id}
In this section, we present a detailed, step-by-step proof of the identifiability of the Average Direct Effect (ADE) and the Average Partial Effect (APE), based on the assumptions outlined in Section \ref{sec:setup}. 

\begin{align}
\tau_{ADE} & = \mathbb{E} \left[\frac{1}{n} \sum_{i \in \mathcal{V}} \tau_{i,DE} \right] \\
& = \mathbb{E} \left[\frac{1}{n} \sum_{i \in \mathcal{V}} Y_i(1,\vect{T_{-i}}) - Y_i(0,\vect{T_{-i}}) \right] = \\
& = \mathbb{E} \left[\frac{1}{n} \sum_{i \in \mathcal{V}} Y_i(1,\vect{T_{\set{N}_i}}) - Y_i(0,\vect{T_{\set{N}_i}}) \right]= \label{eq:id1} \\ 
& = \mathbb{E} \left[\frac{1}{n} \sum_{i \in \mathcal{V}} Y_i(1, z_i) - Y_i(0,z_i) \right] = \label{eq:id2}\\
& = \mathbb{E}_X \left[\mathbb{E} \left[\frac{1}{n} \sum_{i \in \mathcal{V}} Y_i(1, z_i) - Y_i(0,z_i) \mid \matr{X} \right] \right] = \label{eq:id3}\\
& = \mathbb{E}_X \left[\mathbb{E} \left[\frac{1}{n} \sum_{i \in \mathcal{V}} Y_i(1, z_i) - Y_i(0,z_i) \mid \vect{X_i}, \matr{X}_{\set{N}_i} \right] \right] = \label{eq:id4}\\
\begin{split}
& = \mathbb{E}_X \biggl[\mathbb{E} \biggl[\frac{1}{n} \sum_{i \in \mathcal{V}} Y_i(1, z_i) \mid \vect{X_i}, \matr{X}_{\set{N}_i}, t_i, z_i \biggl]  - \\  & \quad \mathbb{E} \biggl[\frac{1}{n} \sum_{i \in \mathcal{V}} Y_i(0,z_i) \mid \vect{X_i}, \matr{X}_{\set{N}_i}, t_i, z_i \biggl] \biggl] 
\end{split}
\label{eq:id5}\\
\begin{split}
& = \mathbb{E}_X \biggl[\mathbb{E} \biggl[\frac{1}{n} \sum_{i \in \mathcal{V}} Y_i \mid \vect{X_i}, \matr{X}_{\set{N}_i}, t_i=1, z_i \biggl]  - \\  & \quad \mathbb{E} \biggl[\frac{1}{n} \sum_{i \in \mathcal{V}} Y_i \mid \vect{X_i}, \matr{X}_{\set{N}_i}, t_i=0, z_i \biggl] \biggl]\label{eq:id6}
\end{split}
\end{align}
Equation \ref{eq:id1} uses partial interference assumption, \ref{eq:id2} uses the assumption that the exposure map is well-defined and known, \ref{eq:id3} uses law of total expectation, \ref{eq:id4} uses partial interference assumption, \ref{eq:id5} uses strong ignorability and \ref{eq:id6} uses consistency assumption.

\begin{align}
\tau_{APE} & = \mathbb{E} \left[\frac{1}{n} \sum_{i \in \mathcal{V}} \tau_{i,PE} \right] \\
& = \mathbb{E} \left[\frac{1}{n} \sum_{i \in \mathcal{V}} Y_i(T_i, \vect{1}_{-i}) - Y_i(T_i, \vect{0}_{-i}) \right] = \\
& = \mathbb{E} \left[\frac{1}{n} \sum_{i \in \mathcal{V}} Y_i(T_i,\vect{T_{\set{N}_i}} = \vect{1}) - Y_i(T_i,\vect{T_{\set{N}_i}}=\vect{0}) \right]= \label{eq:7} \\ 
& = \mathbb{E} \left[\frac{1}{n} \sum_{i \in \mathcal{V}} Y_i(T_i, z_i^\prime) - Y_i(T_i,z_i^\doubleprime) \right] = \label{eq:8}\\
& = \mathbb{E}_X \left[\mathbb{E} \left[\frac{1}{n} \sum_{i \in \mathcal{V}}
Y_i(T_i, z_i^\prime) - Y_i(T_i,z_i^\doubleprime) \mid \matr{X} \right] \right] = \label{eq:9}\\
& = \mathbb{E}_X \left[\mathbb{E} \left[\frac{1}{n} \sum_{i \in \mathcal{V}} Y_i(T_i, z_i^\prime) - Y_i(T_i,z_i^\doubleprime) \mid \vect{X_i}, \matr{X}_{\set{N}_i} \right] \right] = \label{eq:10}\\
\begin{split}
& = \mathbb{E}_X \biggl[\mathbb{E} \biggl[\frac{1}{n} \sum_{i \in \mathcal{V}} Y_i(T_i, z_i^\prime) \mid \vect{X_i}, \matr{X}_{\set{N}_i}, t_i, z_i \biggl]  - \\  & \quad \mathbb{E} \biggl[\frac{1}{n} \sum_{i \in \mathcal{V}} Y_i(T_i,z_i^\doubleprime) \mid \vect{X_i}, \matr{X}_{\set{N}_i}, t_i, z_i \biggl] \biggl] 
\end{split}
\label{eq:11}\\
\begin{split}
& = \mathbb{E}_X \biggl[\mathbb{E} \biggl[\frac{1}{n} \sum_{i \in \mathcal{V}} Y_i \mid \vect{X_i}, \matr{X}_{\set{N}_i}, t_i, z_i=z_i^\prime \biggl]  - \\  & \quad \mathbb{E} \biggl[\frac{1}{n} \sum_{i \in \mathcal{V}} Y_i \mid \vect{X_i}, \matr{X}_{\set{N}_i}, t_i, z_i=z_i^\doubleprime \biggl] \biggl]\label{eq:12}
\end{split}
\end{align}
Equation \ref{eq:7} uses partial interference assumption, \ref{eq:8} uses the assumption that the exposure map is well-defined and known, \ref{eq:9} uses law of total expectation, \ref{eq:10} uses partial interference assumption, \ref{eq:11} uses strong ignorability and \ref{eq:12} uses consistency assumption.

\section{Derivation of score function} \label{score}

In this section, we introduce the concept of the neyman orthogonal score function and proceed to derive the corresponding score function pertinent to our study. This derivation is structured around our specific set of structural equations and is guided by the methodology outlined in ~\citep{chernozhukov2018double}. \\
Let $\zeta \in \mathcal{Z} \subset \mathbb{R}^{d_{\zeta}}$ and $\beta \in 
\mathcal{B} \subset \mathbb{R}^{d_{\beta}}$ be the target and nuisance parameters respectively. Suppose the true parameter values $\zeta_0$ and $\beta_0$ that solves the following optimization problem
$$\max _{\zeta \in \mathcal{Z}, \beta \in \mathcal{B}} \mathbb{E}_P[\mathcal{L}(W ; \zeta, \beta)]$$
where $W$ is a random element taking values in a measurable space $(\mathcal{W},\mathcal{A}_\mathcal{W})$ with law determined by a probability measure $P \in \mathcal{P}_n$ and $\mathcal{L}(W ; \zeta, \beta)$ is a known criterion function. $\zeta_0$ and $\beta_0$ satisfy
$$\mathbb{E}_P\left[\partial_\zeta \mathcal{L}\left(W ; \zeta_0, \beta_0\right)\right]=0, \quad \mathbb{E}_P\left[\partial_\beta \mathcal{L}\left(W ; \zeta_0, \beta_0\right)\right]=0$$

Definition. (neyman orthogonality) The score $\psi=\left(\psi_1, \ldots, \psi_{d_\theta}\right)^{\prime}$ obeys the orthogonality condition at $\left(\theta_0, \eta_0\right)$ with respect to the nuisance realization set $\Gamma_{n} \subset T$ if 
$$\mathbb{E}_P\left[\psi\left(W ; \theta_0, \eta_0\right)\right]=0
$$
holds and the pathwise derivative map $\mathrm{D}_r\left[\eta-\eta_0\right]$ exists for all $r \in[0,1)$ and $\eta \in \Gamma_{n}$ and vanishes at $r=0$; namely,
$$
\partial_\eta \mathbb{E}_P \psi\left(W ; \theta_0, \eta_0\right)\left[\eta-\eta_0\right]=0, \quad \text { for all } \eta \in \Gamma_{n} .
$$
We remark here that the condition holds with $\Gamma_{n}=T$ when $\eta$ is a finite-dimensional vector as long as $\partial_\eta \mathbb{E}_P\left[\psi_j\left(W ; \theta_0, \eta_0\right)\right]=0$ for all $j=1, \ldots, d_\theta$, where $\partial_\eta \mathbb{E}_P\left[\psi_j\left(W ; \theta_0, \eta_0\right)\right]$ denotes the vector of partial derivatives of the function $\eta \mapsto \mathbb{E}_P\left[\psi_j\left(W ; \theta_0, \eta\right)\right]$ for $\eta=\eta_0$.

The neyman orthogonal score function is
$$
\psi(W, \matr{A} ; \zeta, \eta)=\partial_\zeta \mathcal{L}(W ; \zeta, \beta)-\mu \partial_\beta \mathcal{L}(W ; \zeta, \beta)
$$
where $\psi = (\psi_1, ..., \psi_{d_{\zeta}})^\prime$ is a vector of known score functions,the nuisance parameter is
$$
\eta=\left(\beta^{\prime}, \operatorname{vec}(\mu)^{\prime}\right)^{\prime} \in T=\mathcal{B} \times \mathbb{R}^{d_\zeta d_\beta} \subset \mathbb{R}^p, \quad p=d_\beta+d_\zeta d_\beta,
$$
and $\mu$ is the $d_\zeta \times d_\beta$ orthogonalization parameter matrix whose true value $\mu_0$ solves the equation
$$
J_{\zeta \beta}-\mu J_{\beta \beta}=0
$$
for
$$
J=\left(\begin{array}{cc}
J_{\zeta \zeta} & J_{\zeta \beta} \\
J_{\beta \zeta} & J_{\beta \beta}
\end{array}\right)=\left.\partial_{\left(\zeta^{\prime}, \beta^{\prime}\right)} \mathbb{E}_P\left[\partial_{\left(\zeta^{\prime}, \beta^{\prime}\right)^{\prime}} \mathcal{L}(W ; \zeta, \beta)\right]\right|_{\zeta=\zeta_0 ; \beta=\beta_0}
$$

The true value of the nuisance parameter $\eta$ is
$$
\eta_0=\left(\beta_0^{\prime}, \operatorname{vec}\left(\mu_0\right)^{\prime}\right)^{\prime}
$$
and when $J_{\beta \beta}$ is invertible, it has the unique solution,
$$
\mu_0=J_{\zeta \beta} J_{\beta \beta}^{-1}
$$

If $J_{\beta \beta}$ is not invertible, the equation typically has multiple solutions. In this case, it is convenient to focus on a minimal norm solution,
$$
\mu_0=\arg \min \|\mu\| \text { such that }\left\|J_{\zeta \beta}-\mu J_{\beta \beta}\right\|_q=0
$$
for a suitably chosen norm $\|\cdot\|_q$ on the space of $d_\zeta \times d_\beta$ matrices.

In our case, we consider the following criterion function, which is the negative of standard squared loss:
\begin{align*}
       \vect{\mathcal{L}(W ; \zeta, \beta)}_{1 \times 1} = - \frac{\vect{B}^\intercal_{1 \times n} \vect{B}_{n \times 1}}{2}; \quad
    \vect{B}_{n \times 1} = [\vect{Y} - \vect{\ell(\matr{X},\mathcal{A})} - \theta(\vect{T} - \vect{m(\matr{X},\mathcal{A})}) - \alpha(\mathcal{A}(\vect{T} - \vect{m(\matr{X}, \mathcal{A})}))]  
\end{align*}

where $\zeta = (\theta, \alpha)$ are the target parameters and $\beta = (m, \ell)$ are nuisance parameters. $m$ and $\ell$ are estimates of $m_0(\matr{X},\mathcal{A})$
 and $\ell_0(\matr{X},\mathcal{A})$ where $m_0(\matr{X},\mathcal{A}) = \mathbb{E}_P[T|\matr{X},\mathcal{A}]$ and $\ell_0(\matr{X},\mathcal{A}) = \mathbb{E}_P[Y|\matr{X},\mathcal{A}]$. Thus, we want to solve the following maximization problem and find $\theta_0$ and $\alpha_0$ such that
$$\theta_0, \alpha_0 = \arg \max _{\theta \in \Theta, \alpha \in \triangle} \mathbb{E}_P [\vect{\mathcal{L}(W ; \zeta, \beta)}_{1 \times 1}]$$
We take the derivatives to build the score function
\begin{align*}
    \partial_\theta \vect{\mathcal{L}(W ; \zeta, \beta)}_{1 \times 1} &= \vect{B}^\intercal_{1 \times n}(\vect{T} - \vect{m(\matr{X}, \mathcal{A})})_{n \times 1}\\
    \partial_\alpha \vect{\mathcal{L}(W ; \zeta, \beta)}_{1 \times 1} &= \vect{B}^\intercal_{1 \times n} \mathcal{A}(\vect{T} - \vect{m(\matr{X}, \mathcal{A})})_{n \times 1}\\
    \partial_m \vect{\mathcal{L}(W ; \zeta, \beta)}_{1 \times n} &= -\vect{B}^\intercal_{1 \times n}(\theta {I_n} + \alpha \mathcal{A})_{n \times n}\\
    \partial_\ell \vect{\mathcal{L}(W ; \zeta, \beta)}_{1 \times n} &= \vect{B}^\intercal_{1 \times n} {I_n} = \vect{B}^\intercal_{1 \times n}
\end{align*}
$\mathbf{I_n}$ is identity matrix with dimension $n \times n$. 

Let $\vect{B_0}_{n \times 1} = \vect{Y} - \vect{\ell_0(\matr{X},\mathcal{A})} - \theta_0(\vect{T} - \vect{m_0(\matr{X}, \mathcal{A})}) - \alpha_0(\mathcal{A}(\vect{T} - \vect{m_0(\matr{X}, \mathcal{A})}))$
\begin{eqnarray*}
    &&J_{\beta \beta} = \left.\partial_{ \beta^{\prime}} \mathbb{E}_P\left[\partial_{\beta} \vect{\mathcal{L}(W ; \zeta, \beta)}\right]\right|_{\zeta=\zeta_0 ; \beta=\beta_0} \\&& =\left(\begin{array}{cc}
-[(\theta_0{I_n} + \alpha_0 \mathcal{A})^\intercal(\theta_0{I_n} + \alpha_0 \mathcal{A})]_{n \times n} & [(\theta_0{I_n} + \alpha_0 \mathcal{A})^\intercal]_{n \times n}  
\\
\left[(\theta_0 {I_n} + \alpha_0 \mathcal{A})\right]_{n \times n} & -[{I_n}]_{n \times n}
\end{array}\right)_{2n \times 2n} \Rightarrow \text{not invertible}
\end{eqnarray*}

Since $J_{\beta \beta}$ is not invertible, we need to find the minimal norm solution
$$
\mu_0=\arg \min \|\mu\| \text { such that }\left\|J_{\zeta \beta}-\mu J_{\beta \beta}\right\|_q=0
$$
Here $\mu_0$ and $J_{\zeta \beta}$ are $2 \times 2n$ matrices and $J_{\beta \beta}$ is a $2n \times 2n$ matrix.                                                            
\small
\begin{eqnarray*}
&&J_{\zeta \beta}= \left.\partial_{ \zeta^{\prime}} \mathbb{E}_P\left[\partial_{\beta} \vect{\mathcal{L}(W ; \zeta, \beta)}\right]\right|_{\zeta=\zeta_0 ; \beta=\beta_0} = \\&&
\left(\begin{array}{cc}
-\mathbb{E}_P[\vect{B_0}^\intercal + (\vect{m_0(\matr{X},\mathcal{A})}-\vect{T})^\intercal(\theta_0 I_n + \alpha_0 \mathcal{A})]_{1 \times n} & \mathbb{E}_P[(\vect{m_0(\matr{X},\mathcal{A})}-\vect{T})^\intercal]_{1 \times n} \\
 -\mathbb{E}_P[B_0^\intercal \mathcal{A} + (\vect{m_0(\matr{X},\mathcal{A})} - \vect{T})^\intercal \mathcal{A} ^\intercal(\theta_0 I_n + \alpha_0 \mathcal{A})]_{1 \times n}& \mathbb{E}_P[(\vect{m_0(\matr{X},\mathcal{A})}-\vect{T})^\intercal \mathcal{A}^\intercal]_{1 \times n}
\end{array}\right)_{2 \times 2n}
\end{eqnarray*} 
\normalsize
Since $m_0(\matr{X},\mathcal{A}) = \mathbb{E}_P[T|\matr{X},\mathcal{A}]$ and $\ell_0(\matr{X},\mathcal{A}) = \mathbb{E}_P[Y|\matr{X},\mathcal{A}]$, $\mathbb{E}_P[m_0(\matr{X},\mathcal{A}) - T] = 0$ and $\mathbb{E}_P[\ell_0(\matr{X},\mathcal{A}) - Y] = 0$. The expectation of multiplication of a fixed matrix in each of these vectors would also be zero because it would be a linear combination of elements with zero expectation. Thus, $J_{\zeta \beta} = 0$ and by inspection, due to the fact that $\|.\| \geq 0$, $\mu=\mathbf{0}$ would make this norm minimum.

Hence, the score function would be:
$$\psi(\matr{W}, \matr{A} ; \zeta, \eta) = 
\left(\begin{array}{cc}
(\vect{Y} - \vect{\ell(\matr{X},\mathcal{A})} - \theta(\vect{T} - \vect{m(\matr{X}, \mathcal{A})}) - \alpha(\mathcal{A}(\vect{T} - \vect{m(\matr{X}, \mathcal{A})}))^\intercal(\vect{T} - \vect{m(\matr{X}, \mathcal{A})}) \\
(\vect{Y} - \vect{\ell(\matr{X},\mathcal{A})} - \theta(\vect{T} - \vect{m(\matr{X}, \mathcal{A})}) - \alpha(\mathcal{A}(\vect{T} - \vect{m(\matr{X}, \mathcal{A})}))^\intercal \mathcal{A} (\vect{T} - \vect{m(\matr{X}, \mathcal{A})})
\end{array}\right)
$$

\newpage
\enlargethispage{1.5cm}
\section{Proof of Theorem \ref{th: theory} and Corresponding Conditions to Verify} \label{sec: dml_cond}
In this section, we describe the essential regularity conditions and provide their respective proofs. These conditions form the foundational basis for proving Theorem \ref{th: theory}. By demonstrating that the score function fulfills specific assumptions, we can effectively invoke Theorems 3.1 and 3.2, along with Corollary 3.1 from ~\cite{chernozhukov2018double}.\\
This application is crucial for establishing two key properties of our estimators: consistency and asymptotic normality. These 2 are the asymptotic properties of an estimator. Asymptotic refers to a mathematical property of a sequence of random variables or a statistical estimator as the sample size approaches infinity. More specifically, it refers to the behavior of the estimator as the sample size becomes larger and larger. An asymptotic result holds in the limit as the sample size grows infinitely large.\\
 We say that an estimate $\hat{\theta}$ is consistent if $\hat{\theta} \rightarrow \theta_0$ in probability as $n \rightarrow \infty$, where $\theta_0$ is the 'true' unknown parameter of the distribution of the sample.\\
We say that $\hat{\theta}$ is asymptotically normal if
$$
\sqrt{n}\left(\hat{\theta}-\theta_0\right) \xrightarrow{\text{d}} \mathcal{N}\left(0, \sigma_{\theta_0}^2\right)
$$
where $\sigma_{\theta_0}^2$ is called the asymptotic variance of the estimate $\hat{\theta}$. Asymptotic normality says that the estimator not only converges to the unknown parameter, but it converges fast enough, at a rate $1 / \sqrt{n}$, where $n$ is the sample size.

These properties are fundamental in reinforcing the statistical robustness and reliability of our estimators in both finite sample and asymptotic contexts. Besides, they allow us to perform uncertainty quantification and build confidence intervals.\\
The invocation of Theorems 3.1 and 3.2 along with Corollary 3.1 from ~\cite{chernozhukov2018double} are sufficient for proving our Theorem~\ref{th: theory}. \\
In the following discussion, we delve into two distinct sets of conditions as outlined in ~\cite{chernozhukov2018double} that are necessary to invoke these theorems.\\
We use $\|.\|_{P,q}$ to denote the $L^q(P)$ norm, i.e. $\|f\|_{P,q} := \|f(W)\|_{P,q} := (\int \mid 
 f(w) \mid^q dP(w))^{\frac{1}{q}}$.
\enlargethispage{1cm}
\begin{assumption} \label{regconds}
    (Regularity Conditions) Let $c > 0, C > 0, c_1 \geqslant c_0 > 0, q > 4$ and $K \geqslant 2$ be some finite constants; and let $\{\delta_n\}_{n = 1}^{\infty}$ and $\{\Delta_n\}_{n = 1}^{\infty}$ be some sequences of positive constants converging to zero such that $\delta_n \geqslant n_f^{-1/2}$. For all probability laws $P \in \mathcal{P}$ for the triple $W = (\vect{T}, \vect{Y}, \mathcal{X})$ the following
    conditions hold:
    \begin{enumerate}
        \item Equation set \ref{eqn:PLM} holds
        \item $c \leqslant \|\vect{\epsilon^T}\|_{P,2}$ , $ \|\vect{\epsilon^T}\|_{P,q} \leqslant C$, $\leqslant \|\vect{\epsilon^Y}\|_{P,q} \leqslant C$ \label{reg2}
        \item $c \leqslant {\| \vect{\epsilon^Y}^\intercal \vect{\epsilon^T} \|}_{P,2}$, $c \leqslant \mathbb{E}_P\left[\vect{\epsilon^T}^\intercal \vect{\epsilon^T} \right]$, $c \leqslant \mathbb{E}_P\left[\vect{\epsilon^T}^\intercal \mathcal{A}^\intercal \vect{\epsilon^T} \right]$ \label{reg3}
        \item $\|\vect{Y}\|_{P,q} \leqslant C$ \label{Ynorm}
        \item $\vect{\epsilon^T}$ and $ \vect{\epsilon^Y}$ are not eigen vectors of $\mathcal{A}$. \label{reg4}
        \item Given a random subset $I$ of $[n_f]$ of size $n^{\prime}=n_f / K$, the nuisance parameter estimator $\widehat{\eta}_0=\widehat{\eta}_0\left(\left(W_i\right)_{i \in I^c}\right)$ belongs to the realization set $\mathcal{T}_{n}$ with probability at least $1-\Delta_{n}$, where $\eta_0 \in \mathcal{T}_{n}$. \label{C2.1}
        \item Given a random subset $I$ of $[n_f]$ of size $n^{\prime}=n_f / K$, the nuisance parameter estimator $\widehat{\eta}_0=\widehat{\eta}_0\left(\left(W_i\right)_{i \in I^c}\right)$ obeys the following conditions: With $P$-probability no less than $1-\Delta_n$,
$$
\left\|\widehat{\eta}_0-\eta_0\right\|_{P, q} \leqslant C, \quad\left\|\widehat{\eta}_0-\eta_0\right\|_{P, 2} \leqslant \delta_n, \quad \text { and }
$$
for the score $\psi$, where $\widehat{\eta}_0=\left(\widehat{m}_0, \widehat{\ell}_0\right)$,
$$
\left\|\widehat{m}_0-m_0\right\|_{P, 2} \times\left(\left\|\widehat{m}_0-m_0\right\|_{P, 2}+\left\|\widehat{\ell}_0-\ell_0\right\|_{P, 2}\right) \leqslant \delta_n n_f^{-1 / 2} .
$$ \label{reg6}
\end{enumerate}

\end{assumption}
\newpage
\subsection{Condition Set 1: Linear Scores with Approximate Neyman Orthogonality} \label{cond1}
For all $n_f \geqslant 3$ and probability measures $P \in \mathcal{P}_n$ that determines the underlying law of $W$:
\begin{enumerate}
    \item Moment condition vanishes at the true parameter $\zeta_0$: $\mathbb{E}_P\left[\psi\left(W ; \zeta_0, \eta_0\right)\right]=0$
    \item The score function is linear in the sense that: $\psi(W, \matr{A} ; \zeta, \eta) = \psi^a(W, \matr{A} ; \zeta, \eta) \theta + \psi^b(W ; \zeta, \eta) \alpha + \psi^c(W ; \zeta, \eta)$
    \item The map $\eta \rightarrow \mathbb{E}_P\left[\psi\left(W ; \zeta_0, \eta_0\right)\right]$ is twice continuously Gateaux-differentiable.
    \item The score $\psi$ is Neyman orthogonal or, more generally, it is Neyman $\lambda_n$ near-orthogonal at $\left(\zeta_0, \eta_0\right)$ with respect to the nuisance realization set $\Gamma_{n} \subset T$ for
    $$
    \lambda_n:=\sup _{\eta \in \Gamma_{n}}\left\|\partial_\eta \mathbb{E}_P \psi\left(W ; \zeta_0, \eta_0\right)\left[\eta-\eta_0\right]\right\| \leqslant \delta_n n_f^{-1 / 2}
    $$
    \item The identification condition holds; namely, the singular values of the matrix
    $$
    J_{0, a}:=\mathbb{E}_P\left[\psi^a\left(W ; \eta_0\right)\right]
    $$
    are between $c_0$ and $c_1$.
\end{enumerate}
\subsection{Condition Set 2: Score Regularity and Quality of nuisance Parameter Estimators}  \label{cond2}
 For all $n_f \geqslant 3$ and $P \in \mathcal{P}_n$, the following conditions hold:
 \begin{enumerate}
 \item  Given a random subset $I$ of $[n_f]$ of size $n^{\prime}=n_f / K$, the nuisance parameter estimator $\widehat{\eta}_0=\widehat{\eta}_0\left(\left(W_i\right)_{i \in I^c}\right)$ belongs to the realization set $\Gamma_{n}$ with probability at least $1-\Delta_n$, where $\Gamma_{n}$ contains $\eta_0$ and is constrained by the next conditions.
 \item The moment conditions hold: \label{C2.2}
$$
\begin{aligned}
& m_n:=\sup _{\eta \in \Gamma_{n}}\left(\mathbb{E}_P\left[\left\|\psi\left(W ; \zeta_0, \eta\right)\right\|^q\right]\right)^{1 / q} \leqslant c_1, \\
& m_n^{\prime}:=\sup _{\eta \in \Gamma_{n}}\left(\mathbb{E}_P\left[\left\|\psi^a(W ; \eta)\right\|^q\right]\right)^{1 / q} \leqslant c_1 .
\end{aligned}
$$
 \item The following conditions on the statistical rates $r_n, r_n^{\prime}$, and $\lambda_n^{\prime}$ hold: \label{C2.3}
$$
\begin{aligned}
& r_n:=\sup _{\eta \in \Gamma_{n}}\left\|\mathbb{E}_P\left[\psi^a(W ; \eta)\right]-\mathbb{E}_P\left[\psi^a\left(W ; \eta_0\right)\right]\right\| \leqslant \delta_n, \\
& r_n^{\prime}:=\sup _{\eta \in \Gamma_{n}}\left(\mathbb{E}_P\left[\left\|\psi\left(W ; \zeta_0, \eta\right)-\psi\left(W ; \zeta_0, \eta_0\right)\right\|^2\right]\right)^{1 / 2} \leqslant \delta_n, \\
& \lambda_n^{\prime}:=\sup _{r \in(0,1), \eta \in \Gamma_{n}}\left\|\partial_r^2 \mathbb{E}_P\left[\psi\left(W ; \zeta_0, \eta_0+r\left(\eta-\eta_0\right)\right)\right]\right\| \leqslant \delta_n / \sqrt{n_f} .
\end{aligned}
$$
 \item The variance of the score $\psi$ is non-degenerate: All eigenvalues of the matrix
$$
\mathbb{E}_P\left[\psi\left(W ; \zeta_0, \eta_0\right) \psi\left(W ; \zeta_0, \eta_0\right)^{\prime}\right]
$$
are bounded from below by $c_0$.
 \end{enumerate}
In the rest of this section, we attempt to prove the condition sets \ref{cond1} and \ref{cond2} under regularity assumptions \ref{regconds}.
\subsection{Proof of Condition Set 1}
\paragraph{\textbf{C.1.1}}
The true parameter values $\zeta_0$ and $\beta_0$ solve the following optimization problem
$$\max _{\zeta \in \mathcal{Z}, \beta \in \mathcal{B}} \mathbb{E}_P[\mathcal{L}(W ; \zeta, \beta)]$$
where $\mathcal{L}(W ; \zeta, \beta)$ is a known criterion function. $\zeta_0$ and $\beta_0$ satisfy
$$\mathbb{E}_P\left[\partial_\zeta \mathcal{L}\left(W ; \zeta_0, \beta_0\right)\right]=0, \quad \mathbb{E}_P\left[\partial_\beta \mathcal{L}\left(W ; \zeta_0, \beta_0\right)\right]=0$$
The neyman orthogonal score function is
$$
\psi(W, \matr{A} ; \zeta, \eta)=\partial_\zeta \mathcal{L}(W ; \zeta, \beta)-\mu \partial_\beta \mathcal{L}(W ; \zeta, \beta)
$$
Thus, by definition of $\zeta_0$ and $\eta_0$, we have:
$$\mathbb{E}_P\left[\psi\left(W ; \zeta_0, \eta_0\right)\right]=0
$$
\paragraph{\textbf{C.1.2}}
The score function is linear in the sense that:
\scriptsize{
\begin{align}
\psi(W, \matr{A} ; \zeta, \eta) = 
\left(\begin{array}{cc}
(\vect{Y} - \vect{\ell(\matr{X},\mathcal{A})} - \theta(\vect{T} - \vect{m(\matr{X}, \mathcal{A})}) - \alpha(\mathcal{A}(\vect{T} - \vect{m(\matr{X}, \mathcal{A})}))^\intercal(\vect{T} - \vect{m(\matr{X}, \mathcal{A})}) \\
(\vect{Y} - \vect{\ell(\matr{X},\mathcal{A})} - \theta(\vect{T} - \vect{m(\matr{X}, \mathcal{A})}) - \alpha(\mathcal{A}(\vect{T} - \vect{m(\matr{X}, \mathcal{A})}))^\intercal \mathcal{A} (\vect{T} - \vect{m(\matr{X}, \mathcal{A})})
\end{array}\right)
 = \nonumber
 \\
 \nonumber
\underbrace{ \left(\begin{array}{cccc}
-(\vect{T} - \vect{m(\matr{X}, \mathcal{A})})^\intercal(\vect{T} - \vect{m(\matr{X}, \mathcal{A})}) & -(\vect{T} - \vect{m(\matr{X}, \mathcal{A})})^\intercal \mathcal{A}^\intercal(\vect{T} - \vect{m(\matr{X}, \mathcal{A})}) \\
-(\vect{T} - \vect{m(\matr{X}, \mathcal{A})})^\intercal \mathcal{A}(\vect{T} - \vect{m(\matr{X}, \mathcal{A})}) & -(\vect{T} - \vect{m(\matr{X}, \mathcal{A})})^\intercal \mathcal{A}^\intercal \mathcal{A}(\vect{T} - \vect{m(\matr{X}, \mathcal{A})})
\end{array}\right)}_{\psi^a} 
\left(\begin{array}{cc}
\theta \\
\alpha
\end{array}\right)
 + 
\underbrace{ \left(\begin{array}{cc}
(\vect{Y} - \vect{\ell(\matr{X},\mathcal{A})})^\intercal (\vect{T} - \vect{m(\matr{X}, \mathcal{A})}) \\
(\vect{Y} - \vect{\ell(\matr{X},\mathcal{A})})^\intercal \mathcal{A}(\vect{T} - \vect{m(\matr{X}, \mathcal{A})})
\end{array}\right)}_{\psi^b}
\end{align}
}
\normalsize
\paragraph{\textbf{C.1.3}}
The score function can trivially be shown to be twice Gateaux differentiable.
\paragraph{\textbf{C.1.4}}
To show neyman orthogonality, we need to show that Gateaux derivative vanishes in addition to the moment condition.
The Gateaux derivative in the direction $\eta - \eta_0 = (\vect{m} - \vect{m_0}, \vect{\ell} - \vect{\ell_0})$ is: 
\begin{multline*}
\partial_\eta \mathbb{E}_P \psi\left(W ; \zeta_0, \eta_0\right)\left[\eta-\eta_0\right] = 
\\
\\
\scriptsize{\lim\limits_{r \to 0}
\frac{\splitfrac{ \mathbb{E}_P 
\left(\begin{array}{cc}
(\vect{Y} - (\vect{\ell_0}+r(\vect{\ell} - \vect{\ell_0})) - \theta_0(\vect{T} - (\vect{m_0}+r(\vect{m}-\vect{m_0}))) - \alpha_0(\mathcal{A}(\vect{T} - (\vect{m_0}+r(\vect{m}-\vect{m_0})))))^\intercal (\vect{T} - (\vect{m_0}+r(\vect{m}-\vect{m_0}))) \\
(\vect{Y} - (\vect{\ell_0}+r(\vect{\ell} - \vect{\ell_0})) - \theta_0(\vect{T} - (\vect{m_0}+r(\vect{m}-\vect{m_0}))) - \alpha_0(\mathcal{A}(\vect{T} - (\vect{m_0}+r(\vect{m}-\vect{m_0})))))^\intercal \mathcal{A} (\vect{T} - (\vect{m_0}+r(\vect{m}-\vect{m_0})))
\end{array}\right)}{\hspace{-1.5cm} - \mathbb{E}_P 
\left(\begin{array}{cc}
(\vect{Y} - \vect{\ell_0} - \theta_0(\vect{T} - \vect{m_0}) - \alpha_0(\mathcal{A}(\vect{T} - \vect{m_0}))^\intercal(\vect{T} - \vect{m_0}) \\
(\vect{Y} - \vect{\ell_0} - \theta_0(\vect{T} - \vect{m_0}) - \alpha_0(\mathcal{A}(\vect{T} - \vect{m_0}))^\intercal \mathcal{A} (\vect{T} - \vect{m_0})
\end{array}\right)}}{r}} = 
\\
\\
\hspace{-0.5cm}
\scriptsize{
\lim\limits_{r \to 0}
\frac{\splitfrac{\mathbb{E}_P 
\left(\begin{array}{cc}
(\overbrace{\vect{Y} - \vect{\ell_0} - \theta_0(\vect{T} - \vect{m_0}) - \alpha_0(\mathcal{A}(\vect{T} - \vect{m_0}-r(\vect{m}-\vect{m_0})))}^{\vect{\epsilon^Y}}-\overbrace{r(\vect{\ell} - \vect{\ell_0})}^{\vect{G}} + \overbrace{\theta_0 r(\vect{m}-\vect{m_0})}^{\theta_0 \vect{D}}+ \overbrace{\alpha_0 \mathcal{A} r(\vect{m}-\vect{m_0})}^{\alpha_0 \mathcal{A}\vect{D}})^\intercal (\overbrace{\vect{T} - \vect{m_0}}^{\vect{\epsilon^T}}-\overbrace{r(\vect{m}-\vect{m_0})}^{\vect{D}}) \\
(\vect{Y} - \vect{\ell_0} - \theta_0(\vect{T} - \vect{m_0}) - \alpha_0(\mathcal{A}(\vect{T} - \vect{m_0}-r(\vect{m}-\vect{m_0})))-r(\vect{\ell} - \vect{\ell_0}) + \theta_0 r(\vect{m}-\vect{m_0})+\alpha_0 \mathcal{A} r(\vect{m}-\vect{m_0}))^\intercal \mathcal{A} (\vect{T} - \vect{m_0}-r(\vect{m}-\vect{m_0}))
\end{array}\right)}{\hspace{-1.5cm} - \mathbb{E}_P 
\left(\begin{array}{cc}
(\overbrace{{\vect{Y} - \vect{\ell_0} - \theta_0(\vect{T} - \vect{m_0}) - \alpha_0(\mathcal{A}(\vect{T} - \vect{m_0}))}^\intercal}^{\vect{\epsilon^Y}^\intercal} \overbrace{(\vect{T} - \vect{m_0})}^{\vect{\epsilon}^T} \\
(\vect{Y} - \vect{\ell_0} - \theta_0(\vect{T} - \vect{m_0}) - \alpha_0(\mathcal{A}(\vect{T} - \vect{m_0}))^\intercal \mathcal{A} (\vect{T} - \vect{m_0})
\end{array}\right)}}{r}}
\\
\\
=
\scriptsize{
\lim\limits_{r \to 0} \frac{\mathbb{E}_P 
\left( \begin{array}{cc}
 \cancel{{\vect{\epsilon^Y}}^\intercal \vect{\epsilon^T}} - {\vect{\epsilon^Y}}^\intercal \vect{D} - \vect{G}^\intercal \vect{\epsilon^T} +  \theta_0 \vect{D}^\intercal \vect{\epsilon^T} \overbrace{+ \vect{G}\vect{D} -  \theta_0 \vect{D}^\intercal \vect{D} - \alpha_0 \vect{D}^\intercal \mathcal{A}^\intercal \vect{D}}^{\text{goes to 0 includes } r^2} + \alpha_0 \vect{D}^\intercal \mathcal{A}^\intercal \vect{\epsilon^T} - \cancel{{\vect{\epsilon^Y}}^\intercal \vect{\epsilon^T}}\\
\cancel{{\vect{\epsilon^Y}}^\intercal \mathcal{A} \vect{\epsilon^T}} - {\vect{\epsilon^Y}}^\intercal \mathcal{A} \vect{D} - \vect{G}^\intercal 
 \mathcal{A} \vect{\epsilon^T} +  \theta_0 \vect{D}^\intercal \mathcal{A} \vect{\epsilon^T} \overbrace{+ \vect{G}\mathcal{A}\vect{D} -  \theta_0 \vect{D}^\intercal \mathcal{A} \vect{D} - \alpha_0 \vect{D}^\intercal \mathcal{A}^\intercal \mathcal{A} \vect{D}}^{\text{goes to 0 includes } r^2} + \alpha_0 \vect{D}^\intercal \mathcal{A}^\intercal \mathcal{A} \vect{\epsilon^T} - \cancel{{\vect{\epsilon^Y}}^\intercal \mathcal{A} \vect{\epsilon^T}}
\end{array}
\right)}{r}
}
\\
\\
=
\scriptsize{
\lim\limits_{r \to 0} \frac{\cancel{r} \mathbb{E}_P 
\left( \begin{array}{cc}
-{\epsilon^Y}^\intercal(\vect{m}-\vect{m_0}) - (\vect{\ell}-\vect{\ell_0})^\intercal \vect{\epsilon^T} + \theta_0 (\vect{m}-\vect{m_0})^\intercal \vect{\epsilon^T} + \alpha_0 (\vect{m}-\vect{m_0})^\intercal \mathcal{A}^\intercal \vect{\epsilon^T}\\
-{\epsilon^Y}^\intercal \mathcal{A} (\vect{m}-\vect{m_0}) - (\vect{\ell}-\vect{\ell_0})^\intercal \mathcal{A} \vect{\epsilon^T} + \theta_0 (\vect{m}-\vect{m_0})^\intercal \mathcal{A} \vect{\epsilon^T} + \alpha_0 (\vect{m}-\vect{m_0})^\intercal \mathcal{A}^\intercal \mathcal{A} \vect{\epsilon^T}
\end{array}
\right)}{\cancel{r}}
}
\\
\\
=
\scriptsize{
\mathbb{E}_P 
\left( \begin{array}{cc}
-{\epsilon^Y}^\intercal(\vect{m}-\vect{m_0}) - (\vect{\ell}-\vect{\ell_0})^\intercal \vect{\epsilon^T} + \theta_0 (\vect{m}-\vect{m_0})^\intercal \vect{\epsilon^T} + \alpha_0 (\vect{m}-\vect{m_0})^\intercal \mathcal{A}^\intercal \vect{\epsilon^T}\\
-{\epsilon^Y}^\intercal \mathcal{A} (\vect{m}-\vect{m_0}) - (\vect{\ell}-\vect{\ell_0})^\intercal \mathcal{A} \vect{\epsilon^T} + \theta_0 (\vect{m}-\vect{m_0})^\intercal \mathcal{A} \vect{\epsilon^T} + \alpha_0 (\vect{m}-\vect{m_0})^\intercal \mathcal{A}^\intercal \mathcal{A} \vect{\epsilon^T}
\end{array}
\right)}
\end{multline*}
\normalsize
Consider the first term in the above expectation. We use Law of Iterated Expectations:
\small{\begin{align}
\mathbb{E}_P[{\epsilon^Y}^\intercal(\vect{m}-\vect{m_0})] = \mathbb{E}_{X,D,Y}[{\epsilon^Y}^\intercal(\vect{m}-\vect{m_0})] = \mathbb{E}_{X}[\mathbb{E}_{YT\mid X}[{\epsilon^Y}^\intercal \overbrace{(\vect{m}-\vect{m_0})}^{\text{constant given X}} \mid X]]
=\mathbb{E}_{X}[(\vect{m}-\vect{m_0})\mathbb{E}_{YT \mid X}[\overbrace{({\epsilon^Y}^\intercal) \mid X}^0]] = 0 \nonumber
\end{align}}
\normalsize
A similar argument can be used to show that other expectation terms are $0$.
\paragraph{\textbf{C.1.5}}

\begin{eqnarray*}
    &&J_{0, a}:=\mathbb{E}_P\left[\psi^a\left(W ; \eta_0\right)\right]\\&& = \mathbb{E}_P\left[ \left(\begin{array}{cccc}
-(\vect{T} - \vect{m_0(\matr{X}, \mathcal{A})})^\intercal(\vect{T} - \vect{m_0(\matr{X}, \mathcal{A})}) & -(\vect{T} - \vect{m_0(\matr{X}, \mathcal{A})})^\intercal \mathcal{A}^\intercal(\vect{T} - \vect{m_0(\matr{X}, \mathcal{A})}) \\
-(\vect{T} - \vect{m_0(\matr{X}, \mathcal{A})})^\intercal \mathcal{A}(\vect{T} - \vect{m_0(\matr{X}, \mathcal{A})}) & -(\vect{T} - \vect{m_0(\matr{X}, \mathcal{A})})^\intercal \mathcal{A}^\intercal \mathcal{A}(\vect{T} - \vect{m_0(\matr{X}, \mathcal{A})})
\end{array}\right)\right]
\\ &&
= \left(\begin{array}{cccc}
\vect{A_1} & \vect{A_2} \\ \vect{A_3} & \vect{A_4}
\end{array}\right)
\end{eqnarray*}

\begin{align}
    J_{0, a} ^\intercal J_{0, a} &=  \left(\begin{array}{cccc}
\vect{A_1} & \vect{A_3} \\ \vect{A_2} & \vect{A_4}
\end{array}\right)  \left(\begin{array}{cccc}
\vect{A_1} & \vect{A_2} \\ \vect{A_3} & \vect{A_4}
\end{array}\right) = \left(\begin{array}{cccc}
\vect{A_1}^2 + \vect{A_3}^2 & \vect{A_1}\vect{A_2} + \vect{A_3}\vect{A_4} \\ \vect{A_1}\vect{A_2} + \vect{A_3}\vect{A_4} & \vect{A_2}^2 + \vect{A_4}^2
\end{array}\right) \nonumber
\end{align}

The eigen values of this matrix are the roots of the following quadratic equation:
\begin{align}
\lambda^2 - \lambda(\vect{A_1}^2 + \vect{A_2}^2 + \vect{A_3}^2 + \vect{A_4}^2) + (\vect{A_1}^2 + \vect{A_3}^2)(\vect{A_2}^2 + \vect{A_4}^2) - (\vect{A_1}\vect{A_2} + \vect{A_3}\vect{A_4})^2 = 0 \nonumber
\end{align}
We know that in a quadratic equation of form $a_2x^2 + a_1x + a_0 = 0$, the sum of the roots are $\frac{-a_1}{a_2}$ and the product of the roots are $\frac{a_0}{a_2}$. To ensure that all the eigen values are positive, we need to make sure both $\frac{-a_1}{a_2}$ and $\frac{a_0}{a_2}$ are positive:
\begin{align}
& \frac{-a_1}{a_2} = \vect{A_1}^2 + \vect{A_2}^2 + \vect{A_3}^2 + \vect{A_4}^2 \geqslant c > 0 \label{sum_roots} &\\& 
    \frac{a_0}{a_2} =  (\vect{A_1}^2 + \vect{A_3}^2)(\vect{A_2}^2 + \vect{A_4}^2) - (\vect{A_1}\vect{A_2} + \vect{A_3}\vect{A_4})^2 = (A_1A_4 - A_2A_3)^2 \label{prod}
\end{align}
\ref{sum_roots} holds since the summation of squared elements are non-negative and $A_1^2 = \mathbb{E}_P[(\vect{T} - \vect{m_0(\matr{X}, \mathcal{A})})^\intercal(\vect{T} - \vect{m_0(\matr{X}, \mathcal{A})})]^2 = \| \vect{\epsilon^T} \| \geqslant c $  by assumption \ref{reg2}. 
For \ref{prod} to hold, since the squared value is non-negative, we need to show it is not zero, i.e. $A_1 A_4 \neq A_2 A_3$. By Cauchy-Schwarz inequality, we know that:
\begin{align}
\mathbb{E}_P\left[ \vect{\epsilon^T}^\intercal \vect{\epsilon^T} \right] \mathbb{E}_P\left[ \vect{\epsilon^T}^\intercal \mathcal{A}^\intercal \mathcal{A}\vect{\epsilon^T} \right] \geqslant \mathbb{E}_P\left[ \vect{\epsilon^T}^\intercal \mathcal{A}\vect{\epsilon^T} \right] ^2 \nonumber
\end{align}
where the equality holds if $\| \vect{\epsilon^T} \| = 0$ or $\| \mathcal{A}\vect{\epsilon^T} \| = 0$, which does not hold by assumption \ref{reg2} and the fact that $\matr{A}$ is a non-zero matrix. Also the equality can happen if $\exists r: \mathcal{A}\vect{\epsilon^T} = r \vect{\epsilon^T} $, which does not hold by assumption \ref{reg4}. Thus, summation and product of the eigen values are positive, leading to positivity of the singular values of $J_{0, a}$.

Following proposition is derived from ~\citep{gallier2023algebra}:
\begin{proposition} \label{norm_bound}
    For every norm $\|\|$ on $\mathbb{C}^n\left(\right.$ or $\left.\mathbb{R}^n\right)$, for every matrix $A \in \mathrm{M}_n(\mathbb{C})$ (or $A \in \mathrm{M}_n(\mathbb{R})$ ), there is a real constant $C_A \geqslant 0$, such that
$$
\|A u\| \leq C_A\|u\|,
$$
for every vector $u \in \mathbb{C}^n$ (or $u \in \mathbb{R}^n$ if $A$ is real).

\end{proposition}
\ref{norm_bound} states that every linear map on a finite-dimensional space is bounded.

\subsection{Proof of Condition Set 2}

\paragraph{\textbf{C.2.1}}
Condition C.2.1 holds by the construction of the set $\Gamma_{n}$ and Assumption \ref{C2.1}.
\paragraph{\textbf{C.2.2}}
We prove the boundedness of norms of these matrices by showing the bound on the norm of the elements considering the fact that if norm of each element is bounded, then the norm of the matrix is bounded.
We first show the bound for the first elements of $\psi_a$ and $\psi$.:

$$\psi^a(W ; \eta) = 
\left(\begin{array}{cccc}
-(\vect{T} - \vect{m(\matr{X}, \mathcal{A})})^\intercal(\vect{T} - \vect{m(\matr{X}, \mathcal{A})}) & -(\vect{T} - \vect{m(\matr{X}, \mathcal{A})})^\intercal \mathcal{A}^\intercal(\vect{T} - \vect{m(\matr{X}, \mathcal{A})}) \\
-(\vect{T} - \vect{m(\matr{X}, \mathcal{A})})^\intercal \mathcal{A}(\vect{T} - \vect{m(\matr{X}, \mathcal{A})}) & -(\vect{T} - \vect{m(\matr{X}, \mathcal{A})})^\intercal \mathcal{A}^\intercal \mathcal{A}(\vect{T} - \vect{m(\matr{X}, \mathcal{A})})
\end{array}\right)$$

$$\psi(W ; \zeta_0, \eta) = 
\left(\begin{array}{cc}
(\vect{Y} - \vect{\ell(\matr{X},\mathcal{A})} - \theta_0(\vect{T} - \vect{m(\matr{X}, \mathcal{A})}) - \alpha_0(\mathcal{A}(\vect{T} - \vect{m(\matr{X}, \mathcal{A})}))^\intercal(\vect{T} - \vect{m(\matr{X}, \mathcal{A})}) \\
(\vect{Y} - \vect{\ell(\matr{X},\mathcal{A})} - \theta_0(\vect{T} - \vect{m(\matr{X}, \mathcal{A})}) - \alpha_0(\mathcal{A}(\vect{T} - \vect{m(\matr{X}, \mathcal{A})}))^\intercal \mathcal{A} (\vect{T} - \vect{m(\matr{X}, \mathcal{A})})
\end{array}\right)
$$

\begin{align}
    & \left(\mathbb{E}_P\left[\left\|\psi_{11}^a(W ; \eta)\right\|^{q / 2}\right]\right)^{2 / q}=\left\|\psi_{11}^a(W ; \eta)\right\|_{P, q / 2} = \left\| -(\vect{T} - \vect{m(\matr{X}, \mathcal{A})})^\intercal(\vect{T} - \vect{m(\matr{X}, \mathcal{A})} \right\|_{P, q / 2} = & \nonumber\\
    & \left\| -((\vect{T} - \vect{m_0(\matr{X}, \mathcal{A})}) - (\vect{m(\matr{X}, \mathcal{A})} - \vect{m_0(\matr{X}, \mathcal{A})}))^\intercal ((\vect{T} - \vect{m_0(\matr{X}, \mathcal{A})}) - (\vect{m(\matr{X}, \mathcal{A})} - \vect{m_0(\matr{X}, \mathcal{A})})) \right\|_{P, q / 2} = & \nonumber \\
    & \left\| ((\vect{m(\matr{X}, \mathcal{A})} - \vect{m_0(\matr{X}, \mathcal{A})}) - \vect{\epsilon^T})^\intercal (\vect{\epsilon^T} - (\vect{m(\matr{X}, \mathcal{A})} - \vect{m_0(\matr{X}, \mathcal{A})})) \right\|_{P, q / 2} = & \nonumber\\
    & \bigl\| (\vect{m(\matr{X}, \mathcal{A})} - \vect{m_0(\matr{X}, \mathcal{A})})^\intercal \vect{\epsilon^T} - (\vect{m(\matr{X}, \mathcal{A})} - \vect{m_0(\matr{X}, \mathcal{A})})^\intercal(\vect{m(\matr{X}, \mathcal{A})} - \vect{m_0(\matr{X}, \mathcal{A})}) -  \vect{\epsilon^T}^\intercal \vect{\epsilon^T} + \nonumber &\\&  \vect{\epsilon^T}^\intercal (\vect{m(\matr{X}, \mathcal{A})} - \vect{m_0(\matr{X}, \mathcal{A})})\bigr\|_{P, q / 2} \leqslant 
    \left\| \vect{m(\matr{X}, \mathcal{A})} - \vect{m_0(\matr{X}, \mathcal{A})}\right\|_{P, q}\left\| \vect{\epsilon^T}\right\|_{P, q} + \nonumber &\\& 
 \left\| \vect{m(\matr{X}, \mathcal{A})} - \vect{m_0(\matr{X}, \mathcal{A})}\right\|_{P, q}\left\| \vect{m(\matr{X}, \mathcal{A})} - \vect{m_0(\matr{X}, \mathcal{A})} \right\|_{P, q} + \left\| \vect{\epsilon^T}\right\|_{P, q}\left\| \vect{\epsilon^T}\right\|_{P, q} + \nonumber &\\ 
     & \left\| \vect{\epsilon^T}\right\|_{P, q}\left\| \vect{m(\matr{X}, \mathcal{A})} - \vect{m_0(\matr{X}, \mathcal{A})}\right\|_{P, q} \leqslant 4C^2 \nonumber
\end{align}
by assumptions \ref{reg2} and \ref{reg6}.
Following the exact same approach along with proposition \ref{norm_bound}, we can derive an upperbound for other elements of $\left\|\psi^a(W ; \eta)\right\|_{P, q / 2}$, which gives the bound on $m^\prime_n$ in condition \ref{C2.2}.\\
Next, we establish an upper-bound for the first element of $\left(\mathbb{E}_P\left[\left\|\psi(W ; \zeta_0, \eta)\right\|^{q / 2}\right]\right)^{2 / q}$.
First, we need an upper-bound on $\theta_0$ and $\alpha_0$, which will be used later.

\begin{align}
   &\mathbb{E}_P\left[ \psi(W;\zeta_0, \eta_0) \right] = \mathbb{E}_P\left[ \psi^a(W; \eta_0) \right] \left(\begin{array}{cc}
\theta \\
\alpha
\end{array}\right) + \mathbb{E}_P\left[ \psi^b(W; \eta_0) \right] = 0 \nonumber &\\&  
   \theta_0 =  \frac{\mathbb{E}_P\left[ (\vect{Y} - \vect{\ell_0(\matr{X},\mathcal{A})})^\intercal (\vect{m_0(\matr{X}, \mathcal{A})} - \vect{T})\right]}{\mathbb{E}_P\left[ (\vect{T} - \vect{m_0(\matr{X}, \mathcal{A})})^\intercal ( \vect{m_0(\matr{X}, \mathcal{A})} - \vect{T}) \right]} = \frac{\mathbb{E}_P\left[(\vect{Y} - \vect{\ell_0(\matr{X},\mathcal{A})})^\intercal \vect{\epsilon^T}\right]}{\mathbb{E}_P\left[\vect{\epsilon^T}^\intercal \vect{\epsilon^T} \right]}\nonumber &\\&
   \alpha_0 = 
   \frac{\mathbb{E}_P\left[ (\vect{Y} - \vect{\ell_0(\matr{X},\mathcal{A})})^\intercal (\vect{m_0(\matr{X}, \mathcal{A})} - \vect{T})\right]}{\mathbb{E}_P\left[ (\vect{T} - \vect{m_0(\matr{X}, \mathcal{A})})^\intercal \mathcal{A}^\intercal ( \vect{m_0(\matr{X}, \mathcal{A})} - \vect{T}) \right]} =
   \frac{\mathbb{E}_P\left[ (\vect{Y} - \vect{\ell_0(\matr{X},\mathcal{A})})^\intercal \vect{\epsilon^T}\right]}{\mathbb{E}_P\left[ \vect{\epsilon^T}^\intercal \mathcal{A}^\intercal \vect{\epsilon^T} \right]} \nonumber
\end{align}

\begin{align}
 \mid \theta_0 \mid = \frac{\mid \mathbb{E}_P[(\vect{Y} - \vect{\ell_0(\matr{X},\mathcal{A})})^\intercal \vect{\epsilon^T}]\mid}{\mid \mathbb{E}_P\left[\vect{\epsilon^T}^\intercal \vect{\epsilon^T} \right]\mid} \leqslant c^{-1} C (\| Y \|_{P, q} + \| \ell_0(\matr{X},\mathcal{A}) \|_{P, q}) \leqslant 2c^{-1} C (\| Y \|_{P, q}) \leqslant 2C^2/c \nonumber
\end{align}

\begin{align}
 \mid \alpha_0 \mid = \frac{\mid \mathbb{E}_P[(\vect{Y} - \vect{\ell_0(\matr{X},\mathcal{A})})^\intercal \vect{\epsilon^T}]\mid}{\mid \mathbb{E}_P\left[\vect{\epsilon^T}^\intercal \mathcal{A}^\intercal \vect{\epsilon^T} \right]\mid} \leqslant c^{-1} C (\| Y \|_{P, q} + \| \ell_0(\matr{X},\mathcal{A}) \|_{P, q}) \leqslant 2c^{-1} C (\| Y \|_{P, q}) \leqslant 2C^2/c \nonumber
\end{align}

\begin{align}
    & \left(\mathbb{E}_P\left[\left\|\psi_{11}(W ; \zeta_0, \eta)\right\|^{q / 2}\right]\right)^{2 / q}=\left\|\psi_{11}(W ; \zeta_0, \eta)\right\|_{P, q / 2} = \nonumber &\\& \left\|(\vect{Y} - \vect{\ell(\matr{X},\mathcal{A})} - \theta_0(\vect{T} - \vect{m(\matr{X}, \mathcal{A})}) - \alpha_0(\mathcal{A}(\vect{T} - \vect{m(\matr{X}, \mathcal{A})}))^\intercal(\vect{T} - \vect{m(\matr{X}, \mathcal{A})}) \right\|_{P, q / 2} = \nonumber & \\ &
    \bigl|\bigl|(\vect{Y} - \vect{\ell_0(\matr{X},\mathcal{A})} - (\vect{\ell(\matr{X},\mathcal{A})} - \vect{\ell_0(\matr{X},\mathcal{A})}) - \theta_0((\vect{T} - \vect{m_0(\matr{X}, \mathcal{A})}) - (\vect{m(\matr{X}, \mathcal{A})} - \vect{m_0(\matr{X}, \mathcal{A})})) - \nonumber &\\& \alpha_0(\mathcal{A}((\vect{T} - \vect{m_0(\matr{X}, \mathcal{A})}) - (\vect{m(\matr{X}, \mathcal{A})} - \vect{m_0(\matr{X}, \mathcal{A})}))))^\intercal ((\vect{T} - \vect{m_0(\matr{X}, \mathcal{A})}) - (\vect{m(\matr{X}, \mathcal{A})} - \vect{m_0(\matr{X}, \mathcal{A})})) \bigr|\bigr|_{P, q / 2} = \nonumber &\\&
    \bigl\| (\vect{\epsilon^Y} - (\vect{\ell(\matr{X},\mathcal{A})} - \vect{\ell_0(\matr{X},\mathcal{A})}) + \theta_0(\vect{m(\matr{X}, \mathcal{A})} - \vect{m_0(\matr{X}, \mathcal{A})}) + \nonumber &\\& \alpha_0 \mathcal{A}(\vect{m(\matr{X}, \mathcal{A})} - \vect{m_0(\matr{X}, \mathcal{A})}))^\intercal(\vect{\epsilon^T} - (\vect{m(\matr{X}, \mathcal{A})} - \vect{m_0(\matr{X}, \mathcal{A})})) \bigr\| \leqslant \nonumber &\\& (2C + 2C^3/c + 2C_AC^3/c)2C = 4C^2 + 4C^4/c + 4C_AC^4/c \nonumber
\end{align}
where $C_A$ is the constant term introduced in proposition \ref{norm_bound}, which gives the bound on $m_n$ in condition \ref{C2.2}. Following the exact same approach along with proposition \ref{norm_bound}, we can derive an upperbound for other elements of $\left\|\psi(W ; \zeta_0, \eta)\right\|_{P, q / 2}$.

\paragraph{\textbf{C.2.3}}
Following the same argument in the previous section, we prove the boundedness of elements of these matrices:

\begin{align}
    &\left\|\mathrm{E}_P\left[\psi_{11}^a(W ; \eta)\right]-\mathrm{E}_P\left[\psi_{11}^a\left(W ; \eta_0\right)\right]\right\|=\left|\mathrm{E}_P\left[\psi_{11}^a(W ; \eta)-\psi_{11}^a\left(W ; \eta_0\right)\right]\right| = \nonumber&\\&
    \left| \mathrm{E}_P\left[-(\vect{T} - \vect{m(\matr{X}, \mathcal{A})})^\intercal(\vect{T} - \vect{m(\matr{X}, \mathcal{A})}) + (\vect{T} - \vect{m_0(\matr{X}, \mathcal{A})})^\intercal(\vect{T} - \vect{m_0(\matr{X}, \mathcal{A})})\right]\right| = \nonumber&\\&
    \bigl| \mathrm{E}_P\bigl[-((\vect{T} - \vect{m_0(\matr{X}, \mathcal{A})}) - (\vect{m(\matr{X}, \mathcal{A})} - \vect{m_0(\matr{X}, \mathcal{A})}))^\intercal((\vect{T} - \vect{m_0(\matr{X}, \mathcal{A})}) - (\vect{m(\matr{X}, \mathcal{A})} - \vect{m_0(\matr{X}, \mathcal{A})}))  + \nonumber&\\& 
    (\vect{T} - \vect{m_0(\matr{X}, \mathcal{A})})^\intercal(\vect{T} - \vect{m_0(\matr{X}, \mathcal{A})})\bigr] \bigr| = \bigl| \mathrm{E}_P\bigl[((\vect{m(\matr{X}, \mathcal{A})} - \vect{m_0(\matr{X}, \mathcal{A})}) - \vect{\epsilon^T})^\intercal(\vect{\epsilon^T} - (\vect{m(\matr{X}, \mathcal{A})} - \vect{m_0(\matr{X}, \mathcal{A})}))  +  \nonumber&\\&
    \vect{\epsilon^T}^\intercal\vect{\epsilon^T}\bigr] \bigr| = 
    \bigl| \mathrm{E}_P\bigl[(\vect{m(\matr{X}, \mathcal{A})} - \vect{m_0(\matr{X}, \mathcal{A})})^\intercal \vect{\epsilon^T} + \vect{\epsilon^T}^\intercal (\vect{m(\matr{X}, \mathcal{A})} - \vect{m_0(\matr{X}, \mathcal{A})}) - \nonumber&\\&(\vect{m(\matr{X}, \mathcal{A})} - \vect{m_0(\matr{X}, \mathcal{A})})^\intercal 
    (\vect{m(\matr{X}, \mathcal{A})} - \vect{m_0(\matr{X}, \mathcal{A})})\bigr] \bigr| = 2\left\| \vect{\epsilon^T} \right\|_{P,2} \left\| \vect{m(\matr{X}, \mathcal{A})} - \vect{m_0(\matr{X}, \mathcal{A})} \right\|_{P,2} + \nonumber&\\& \left\| \vect{m(\matr{X}, \mathcal{A})} - \vect{m_0(\matr{X}, \mathcal{A})} \right\|^2_{P,2} \leqslant 2C \delta_n + \delta^2_n \leqslant \delta^{\prime}_n \nonumber
\end{align}
by assumption \ref{reg6}, which gives the bound on $r_n$ in condition \ref{C2.3}. Further,
\begin{align}
&\left(\mathrm{E}_P\left[\left\|\psi_{11}\left(W ; \zeta_0, \eta\right)-\psi_{11}\left(W ; \zeta_0, \eta_0\right)\right\|^2\right]\right)^{1 / 2} =\left\|\psi_{11}\left(W ; \theta_0, \eta\right)-\psi_{11}\left(W ; \theta_0, \eta_0\right)\right\|_{P, 2} =  \nonumber &\\& \bigl\| (\vect{Y} - \vect{\ell(\matr{X},\mathcal{A})} - \theta_0(\vect{T} - \vect{m(\matr{X}, \mathcal{A})}) - \alpha_0(\mathcal{A}(\vect{T} - \vect{m(\matr{X}, \mathcal{A})}))^\intercal(\vect{T} - \vect{m(\matr{X}, \mathcal{A})})) - ((\vect{Y} - \vect{\ell_0(\matr{X},\mathcal{A})} - \nonumber &\\& 
\theta_0(\vect{T} - \vect{m_0(\matr{X}, \mathcal{A})}) - \alpha_0(\mathcal{A}(\vect{T} - \vect{m_0(\matr{X}, \mathcal{A})}))^\intercal(\vect{T} - \vect{m_0(\matr{X}, \mathcal{A})}))
)\bigr\|_{P, 2} = \nonumber &\\&
 \bigl\|-\vect{\epsilon^Y}^\intercal (\vect{m(\matr{X}, \mathcal{A})} - \vect{m_0(\matr{X}, \mathcal{A})})  + (\vect{\ell_0(\matr{X},\mathcal{A})} - \vect{\ell(\matr{X},\mathcal{A})} + \theta_0(\vect{m(\matr{X}, \mathcal{A})} - \vect{m_0(\matr{X}, \mathcal{A})})  + \nonumber &\\& \alpha_0 \mathcal{A} (\vect{m(\matr{X}, \mathcal{A})} - \vect{m_0(\matr{X}, \mathcal{A})})
)^\intercal \vect{\epsilon^T} - 
(\vect{\ell_0(\matr{X},\mathcal{A})} - \vect{\ell(\matr{X},\mathcal{A})} + \theta_0(\vect{m(\matr{X}, \mathcal{A})} - \vect{m_0(\matr{X}, \mathcal{A})})  + \nonumber &\\& \alpha_0 \mathcal{A} (\vect{m(\matr{X}, \mathcal{A})} - \vect{m_0(\matr{X}, \mathcal{A})})
)^\intercal (\vect{m(\matr{X}, \mathcal{A})} - \vect{m_0(\matr{X}, \mathcal{A})})
\bigr\|_{P, 2} \leqslant \nonumber &\\&
(C + 2C^3/c + 2C_AC^3/c) \bigl\|  \vect{m(\matr{X}, \mathcal{A})} - \vect{m_0(\matr{X}, \mathcal{A})} \bigr\| + C \bigl\| \vect{\ell(\matr{X},\mathcal{A})} - \vect{\ell_0(\matr{X},\mathcal{A})} \bigr\| + \nonumber &\\&
(\bigl\| \vect{\ell(\matr{X},\mathcal{A})} - \vect{\ell_0(\matr{X},\mathcal{A})} \bigr\| + (2C^2/c + 2C_AC^2/c)\bigl\| \vect{m(\matr{X},\mathcal{A})} - \vect{m_0(\matr{X},\mathcal{A})} \bigr\|) \bigl\|  \vect{m(\matr{X}, \mathcal{A})} - \vect{m_0(\matr{X}, \mathcal{A})} \bigr\| \leqslant \nonumber &\\&
 (1 + 2C^2/c + 2C_AC^2/c)\delta_n n_f^{-1 / 2} \leqslant (1 + 2C^2/c + 2C_AC^2/c)\delta_n \leqslant \delta^{\prime}_n \nonumber
\end{align}
by assumption \ref{reg6}. Following the same approach along with proposition \ref{norm_bound}, we can derive an upper bound for the other dimensions of $\psi$ and $\psi^a$. This upper bound provides the bound on $r_n$ in condition \ref{C2.3}.\\
Lastly, let
\begin{align}
    f(r):=\mathbb{E}_P\left[\psi\left(W ; \theta_0, \eta_0+r\left(\eta-\eta_0\right)\right], \quad r \in(0,1) .\right. \nonumber
\end{align}

Then for any $r \in(0,1)$, for the first dimension of the score function:
\begin{align*}
    &f(r)=\mathbb{E}_P
    \begin{pmatrix}
        (\vect{Y} - (\vect{\ell} + r(\vect{\ell} - \vect{\ell_0})) - \theta_0 (\vect{T} - (\vect{m} + r (\vect{m} - \vect{m_0})))\\ - \alpha_0 \mathcal{A} (\vect{T} - (\vect{m} + r(\vect{m} - \vect{m_0})))) ^\intercal (\vect{T} - (\vect{m} + r(\vect{m} - \vect{m_0})))
    \end{pmatrix} &\\&
    \partial f(r) = \mathbb{E}_P\bigl[(\vect{\ell_0} - \vect{\ell} + \theta_0(\vect{m} - \vect{m_0}) + \alpha_0 \mathcal{A} (\vect{m} - \vect{m_0}))^\intercal (\vect{T} - \vect{m} - r(\vect{m} - \vect{m_0})) + &\\&
    (\vect{Y} - \vect{\ell} - r(\vect{\ell} - \vect{\ell_0}) - \theta_0(\vect{T} - \vect{m} - r(\vect{m} - \vect{m_0})) - \alpha_0 \mathcal{A} (\vect{T} - \vect{m} - r(\vect{m} - \vect{m_0})) )^\intercal(\vect{m_0} - \vect{m}) \bigr] &\\&
    \partial^2 f(r) = \mathbb{E}_P \begin{pmatrix}
        (\vect{\ell_0} - \vect{\ell} + \theta_0(\vect{m} - \vect{m_0}) + \alpha_0 \mathcal{A} (\vect{m} - \vect{m_0}))^\intercal (\vect{m_0} - \vect{m})\\ \quad + (\vect{\ell_0} - \vect{\ell} + \theta_0(\vect{m} - \vect{m_0}) + \alpha_0 \mathcal{A} (\vect{m} - \vect{m_0}))^\intercal (\vect{m_0} - \vect{m}) 
    \end{pmatrix}&\\& = 
        2\mathbb{E}_P\bigl[(\vect{\ell_0} - \vect{\ell} + \theta_0(\vect{m} - \vect{m_0}) + \alpha_0 \mathcal{A} (\vect{m} - \vect{m_0}))^\intercal (\vect{m_0} - \vect{m})  \bigr]&\\&  \leqslant 2 (\| \ell - \ell_0 \| + 2C^2/c\| m - m_0 \| + 2C_AC^2/c\| m - m_0 \|) \|\vect{m} - \vect{m_0}\| &\\& \leqslant 2(1 + 2C^2/c + 2C_AC^2/c)\delta_n n_f^{-1 / 2} \leqslant \delta^\prime_n n_f^{-1 / 2}
\end{align*}
which gives the bound on $\lambda^\prime_n$ in condition \ref{C2.3}.
\paragraph{\textbf{C.2.4}}

\begin{align*}\psi(W ; \zeta_0, \eta_0) = 
\left(\begin{array}{cc}
(\vect{Y} - \vect{\ell_0(\matr{X},\mathcal{A})} - \theta_0(\vect{T} - \vect{m_0(\matr{X}, \mathcal{A})}) - \alpha_0(\mathcal{A}(\vect{T} - \vect{m_0(\matr{X}, \mathcal{A})}))^\intercal(\vect{T} - \vect{m_0(\matr{X}, \mathcal{A})}) \\
(\vect{Y} - \vect{\ell_0(\matr{X},\mathcal{A})} - \theta_0(\vect{T} - \vect{m_0(\matr{X}, \mathcal{A})}) - \alpha_0(\mathcal{A}(\vect{T} - \vect{m_0(\matr{X}, \mathcal{A})}))^\intercal \mathcal{A} (\vect{T} - \vect{m_0(\matr{X}, \mathcal{A})})
\end{array}\right)\\ = 
\left(\begin{array}{cc}
\vect{\epsilon^Y}^\intercal \vect{\epsilon^T} \\
\vect{\epsilon^Y}^\intercal  \mathcal{A} \vect{\epsilon^T}
\end{array}\right)
\end{align*}
\begin{align} \label{var}
\mathbb{E}_P\left[\psi\left(W ; \zeta_0, \eta_0\right) \psi\left(W ; \theta_0, \eta_0\right)^{\prime}\right] =
\mathbb{E}_P\left[
\left(\begin{array}{cc}
\vect{\epsilon^Y}^\intercal \vect{\epsilon^T} \\
\vect{\epsilon^Y}^\intercal  \mathcal{A} \vect{\epsilon^T}
\end{array}\right)
\left(\begin{array}{cc}
\vect{\epsilon^Y}^\intercal \vect{\epsilon^T} &
\vect{\epsilon^Y}^\intercal  \mathcal{A} \vect{\epsilon^T}
\end{array}\right)
\right] \nonumber \\ = \mathbb{E}_P\left[ \left(\begin{array}{cc}
(\vect{\epsilon^Y}^\intercal \vect{\epsilon^T})^2 &
(\vect{\epsilon^Y}^\intercal \vect{\epsilon^T})(\vect{\epsilon^Y}^\intercal  \mathcal{A} \vect{\epsilon^T})\\
(\vect{\epsilon^Y}^\intercal  \mathcal{A} \vect{\epsilon^T})(\vect{\epsilon^Y}^\intercal \vect{\epsilon^T}) &
(\vect{\epsilon^Y}^\intercal  \mathcal{A} \vect{\epsilon^T})^2
\end{array}\right) \right]
\end{align}
The eigen values of this matrix are the roots of the following quadratic equation:
\begin{align} \label{quad2}
\lambda^2 - \lambda(\mathbb{E}_P[ (\vect{\epsilon^Y}^\intercal \vect{\epsilon^T})^2 + (\vect{\epsilon^Y}^\intercal \mathcal{A} \vect{\epsilon^T})^2]) + \mathbb{E}_P[(\vect{\epsilon^Y}^\intercal \vect{\epsilon^T})^2] + \mathbb{E}_P[(\vect{\epsilon^Y}^\intercal \mathcal{A} \vect{\epsilon^T})^2] - \mathbb{E}_P[(\vect{\epsilon^Y}^\intercal \vect{\epsilon^T})(\vect{\epsilon^Y}^\intercal \mathcal{A} \vect{\epsilon^T})]^2 = 0
\end{align}
We know that in a quadratic equation of form $a_2x^2 + a_1x + a_0 = 0$, the sum of the roots are $\frac{-a_1}{a_2}$ and the product of the roots are $\frac{a_0}{a_2}$. To ensure that all the eigen values are positive, we need to make sure both $\frac{-a_1}{a_2}$ and $\frac{a_0}{a_2}$ are positive:
\begin{align}
& \frac{-a_1}{a_2} = \| \vect{\epsilon^Y}^\intercal \vect{\epsilon^T} \|_{P,2} + \| \vect{\epsilon^Y}^\intercal \mathcal{A} \vect{\epsilon^T} \|_{P,2} > 0 \label{sum} &\\& 
    \frac{a_0}{a_2} =  \mathbb{E}_P[(\vect{\epsilon^Y}^\intercal \vect{\epsilon^T})^2] + \mathbb{E}_P[(\vect{\epsilon^Y}^\intercal \mathcal{A} \vect{\epsilon^T})^2] - \mathbb{E}_P[(\vect{\epsilon^Y}^\intercal \vect{\epsilon^T})(\vect{\epsilon^Y}^\intercal \mathcal{A} \vect{\epsilon^T})]^2 > 0 \label{prod2}
\end{align}
\ref{sum} holds according to assumption \ref{reg3}. Equation \ref{prod2} also holds according to Cauchy-Schwarz inequality. The equality in Cauchy-Schwarz inequality for two random variables $X$ and $Y$ happens when $\|X\| = 0$ or $\|Y\| = 0$ or $Y = rX$ for some $r \neq 0$. neither of these cases hold: $\| \vect{\epsilon^Y}^\intercal \vect{\epsilon^T} \|_{P,2} > 0$ based on \ref{reg3}. $\| \vect{\epsilon^Y}^\intercal \mathcal{A} \vect{\epsilon^T} \|_{P,2} > 0$ based on \ref{reg3} and the fact that $\mathcal{A}$ is the adjacency matrix with non-negative elements and $\mathcal{A} \neq 0$. Also, $\nexists r \neq 0: \vect{\epsilon^Y}^\intercal \mathcal{A} \vect{\epsilon^T} = r \vect{\epsilon^Y}^\intercal \vect{\epsilon^T}$ according to \ref{reg4}.
Thus, the roots of equation \ref{quad2}, which are the eigen values of matrix \ref{var} are bounded from below by some positive $c_0$.\\

Thus, all conditions \ref{cond1} and \ref{cond2} are verified. This completes the proof.

\section{Complementary Experimental Setup} \label{exp_appendix}
\subsection{Datasets Details} \label{dataset_appendix}
 We use the following network datasets for our evaluations:
\begin{itemize}
   \item Real World Data
   \begin{itemize}
   \item IndianVillage~\citep{banerjee2014gossip, jackson2012social}: It is a 2010 survey data from villages in Karnataka, India. The survey gathered information from 16,995 individuals residing in 77 villages. It includes 15 features like age, occupation, gender, and more. Additionally, the dataset incorporated 12 distinct social networks involving 69,000 individuals, which included both the surveyed group of 16,995 individuals and others. These networks represented relationships like friendships, relatives, social visits, and financial exchanges. We treated all these connections uniformly, ensuring a consistent network where all edges carried the same meaning.
   \end{itemize}
   \item Semi-Synthetic Data\\ 
   $(\mathcal{X},\vect{T},\vect{Y})$ are generated based on data generative process \ref{dgp} and the network comes from real-world network dataset below:
    \begin{itemize}
    \item Cora~\citep{mccallum2000automating}: It comprises academic research papers and their citation links, forming a graph structure. It consists of 2708 scientific publications classified into one of seven classes. The citation network consists of 5429 links.
    \item Pubmed: Similar to Cora, it is a citation network, consists of 19717 scientific publications from PubMed database classified into one of three classes. The citation network consists of 44338 links.
    \item Flickr: It is a network derived from Flickr, one of the largest platform for sharing photos. Each node in the graph represents an image, and if two images have shared characteristics like geographic location, gallery, or comments by the same user, there will be an edge connecting their respective nodes. It consists of 105938 nodes and 2316948 edges.
    \end{itemize}
    \item Synthetic Data\\
    $(\mathcal{X},\vect{T},\vect{Y})$ are generated based on data generative process \ref{dgp} and the network comes from the synthetic network generative process below:
    \begin{itemize}
    \item Stochastic Block Model (SBM) ~\citep{holland1983stochastic}: It is a generative model for networks. We also tried our method on a synthetic network produced by SBM, to have more control over the network parameters.  In SBM, nodes are partitioned into multiple blocks or communities, and the probability of an edge existing between two nodes depends on their respective block assignments. 
    \end{itemize}
\end{itemize}

\subsection{Data Generative Process} \label{dgp_appendix}
The covariates $(\mathcal{X})$, treatment assignments $(\vect{T})$, and outcomes $(\vect{Y})$ are synthetically generated following a specific data generative process outlined in Section \ref{sec:setup}. This section details one such data-generative process used in our experiments:

\begin{align}
    \mathcal{X} &\sim \mathcal{N}(0,1) \nonumber \\
    \vect{\pi} &= \left(1 + \exp(\frac{\mathcal{X} + \gamma \mathcal{A}\mathcal{X})}{-10}\right)^{-1} \nonumber \\
    \vect{T} &\sim \text{Bin}(\vect{\pi}) \nonumber \\
    \vect{Y} &= \mathcal{X} + \mathcal{A}\mathcal{X} + \vect{T} \times \theta_0 + \alpha \mathcal{A}\vect{T} \label{dgp1}
\end{align}

Where $\mathcal{A}\vect{T}$ is the exposure map and would be the sum of treated neighbors for each node.
In our setup, we assumed that this exposure map is known. In the experiments in which we compare our method against baselines, the target parameters are $\theta_0 = 10$ and $\alpha_0 = 5$. For the Pubmed and Flickr datasets, we adopt this data generative process to generate $(\mathcal{X}, \vect{T}, \vect{Y})$.

We also simulated another data generative process, which is more complex and involves non-linearity:
\begin{align} 
    \vect{X_i} &\sim \mathcal{N}(0,1) \nonumber \\
    \vect{\pi} &= \left(1 + \exp(\frac{\mathcal{X} + \gamma \mathcal{A}\mathcal{X})}{-10}\right)^{-1} \nonumber \\
    T_i &\sim \text{Bin}(\pi_i) \nonumber \\
    Y_i &= \sigma(\sum_j X_{ij} + MAX_j((A\mathcal{X})_{ij})) + T_i \times \theta_0 + \alpha \vect{\mathcal{A}_i}\vect{T} \label{dgp2}
\end{align}
where $\sigma$ denotes the sigmoid function. The target parameters are $\theta_0 = 20$ and $\alpha_0 = 5$.
For the Cora dataset, we adopt this data generative process to generate $(\mathcal{X}, \vect{T}, \vect{Y})$.

\begin{figure*}[t]
    \centering
    \includegraphics[width=1.0\textwidth]{figures/comparison.pdf}
    \caption{Relative Error of different methods for estimating causal effects across different datasets. Note that the y-axis is log-scaled. In the figure, two variants of our method are presented: one utilizing a focal set and another without a focal set, encompassing the entire dataset. 'PA' refers to Double Machine Learning combined with predefined aggregates}
    \label{fig:baselines}
\end{figure*}

\subsection{Extended Baselines} 
\label{ex:baselines}
This section provides a more thorough description of the baselines used.

1. \textbf{NetEst} \citep{jiang2022estimating}: Utilizes GNNs for learning representations of confounders for individual units and their neighbors, coupled with an adversarial learning process to align distributions for networked causal inference.

2. \textbf{Net-TMLE} \citep{ogburn2022causal}: Employs an efficient influence function and moment condition to derive a doubly robust estimator. This approach leverages the efficiency of targeted maximum likelihood estimation (TMLE) to improve the robustness and accuracy of causal effect estimates in the presence of network interference.

3. \textbf{T-Learner} \citep{tlearner}: Creates two separate models to predict outcomes for each treatment arm based on unit and neighbor covariates, with estimations modeled using GNNs. This method provides a straightforward way to estimate treatment effects by splitting the problem into two learning tasks.

4. \textbf{DML with predefined aggregates}: Applies Double Machine Learning (DML) in the i.i.d. setting but uses predefined aggregates like max, min, and mean for neighbor information aggregation. This approach simplifies the network structure into summary statistics, facilitating the application of traditional DML techniques.

5. \textbf{Tresp \& Ma} \citep{Ma2020CausalIU}: Maps the representation of covariates to a new space where treatment and covariates are disentangled, incorporating the Hilbert-Schmidt Independence Criterion (HSIC) as a regularization term. Subsequently, GNNs are employed to aggregate covariate information from neighboring nodes. Two separate models are then trained to estimate the outcome based on the output of the GNNs and a predefined exposure map for the treatment and control groups.

6. \textbf{L\&L method} \citep{leung2022unconfoundedness}.In our paper, we designate this approach as the "L\&L" method in the experiment section. L\&L is a working paper and the code is not publicly available and according to the authors, will become available after the work is published. We implemented a version of their method to the best of our understanding. Unlike methods that utilize continuous exposure measures, the L\&L method requires the conversion of exposure data into a binary format for application to our dataset. Specifically, for each node, if more than half of its neighbors, including the node itself, are subject to treatment, we assign an exposure value of 1; if not, the exposure value is set to 0. Utilizes a standard doubly robust estimator combined with GNNs to estimate the total effect.
\section{Complementary Experimental Results}\label{sec: additional_exps}
\subsection{Coverage Study}
In our investigation, we performed a comprehensive coverage analysis leveraging the closed-form formula for calculating variance and confidence intervals as detailed in Section~\ref{th: theory}. This analysis involved applying our proposed methodology across multiple executions—specifically, 100 iterations—on each dataset under consideration. For each iteration, we computed confidence intervals and assessed the frequency at which the true value of the target parameter fell within these intervals. This measure of frequency serves as a critical indicator of the reliability and precision of our methodology in capturing the parameter of interest across varied datasets. The  results are presented in table~\ref{table:coverage}.
\begin{table}
\vskip 0.15in
\begin{center}
\begin{tabular}{lcccr}
\toprule
    \textbf{Dataset} & \textbf{$\theta$} & \textbf{$\alpha$} \\
\midrule
     Cora & 100\% & 100\%\\
     Pubmed & 100\% & 100\%\\
     Flickr & 92\% &  52\%\\
\bottomrule
\end{tabular}
\end{center}
  \caption{The results of coverage study our approach across datasets over 100 trials with the 95\% confidence interval}
  \label{table:coverage}
\end{table}

\subsection{Case Study: SHG Participation} \label{casestudy}
Below is the table presenting the results of six baseline methods and two versions of our framework on the Indian Village dataset:
\begin{table}[H]
\begin{center}
\resizebox{\columnwidth}{!}{
\begin{tabular}{lccccccccr}
\toprule
    \textbf{} & \textbf{GDML} & \textbf{w/o Focal Set} & \textbf{PA}     & \textbf{T-learner} & \textbf{NetEst}         & \textbf{Net TMLE} & \textbf{L\&L} & \textbf{Ma \& Tresp} \\
\midrule
     ADE & 0.315 & 0.390 & 0.209 & 0.469 & N/A* & 0.291 & N/A & 0.295\\
     APE & 0.050 & -0.002 & -0.004 & N/A & N/A* & N/A & N/A & 0.016\\
     ATE & 0.365 & 0.388 & 0.205 & N/A & N/A* & N/A & 0.113 & 0.311\\
\bottomrule
\end{tabular}}
\end{center}
  \caption{The results of six baseline methods and two versions of our framework on the Indian Village dataset}
  \label{table:real}
\end{table}
NetEst was not stable during training on this data and resulted in NaN values. Note that the confidence intervals for ADE and APE generated by our framework are $[-1.570, 2.200]$ and $[-1.017, 1.116]$ respectively with a 95\% confidence. In the literature, ~\citep{gilad2021heterogeneous} also quantified the direct effect of SHG membership on the probability of possessing an outstanding loan as \textbf{0.30}.

\subsection{Graph density} \label{density_appendix}
 To assess the impact of network data sparsity on estimation performance, we utilize Stochastic Block Model (SBM) synthetic graphs, providing greater control over graph generation.  \\
 We fix the number of components as $200$, the number of nodes as $3000$, and the probability of existence of an edge between components as $0.0001$. Subsequently, we vary the probability of edge existence within the component, denoted as $P_{intra}$, to modulate the sparsity of the graph. For each $P_{intra}$, we generate a single graph and for each graph, we generate $100$ different datasets $\mathcal{X}, \vect{T}$ and $\vect{Y}$ and report the average of estimated direct effect. Table \ref{table:SBM} presents the results. notably, as $P_{intra}$ increases, the number of edges rises. Given the fixed number of nodes, this causes a reduction in the size of the focal set (sample size), resulting in an increased bias in the estimation process. 
This result showcases that our methodology exhibits enhanced performance in sparser networks. As the number of edges increases within a network with a fixed number of nodes, we observe a corresponding rise in both the relative error and the variance of our estimations. This trend suggests a direct relationship between network density and the performance of our method.

\captionsetup[table]{skip=10pt}
\begin{figure}[H]
\hspace*{-0.5cm}  
    \centering
    \includegraphics[width=0.55\textwidth]{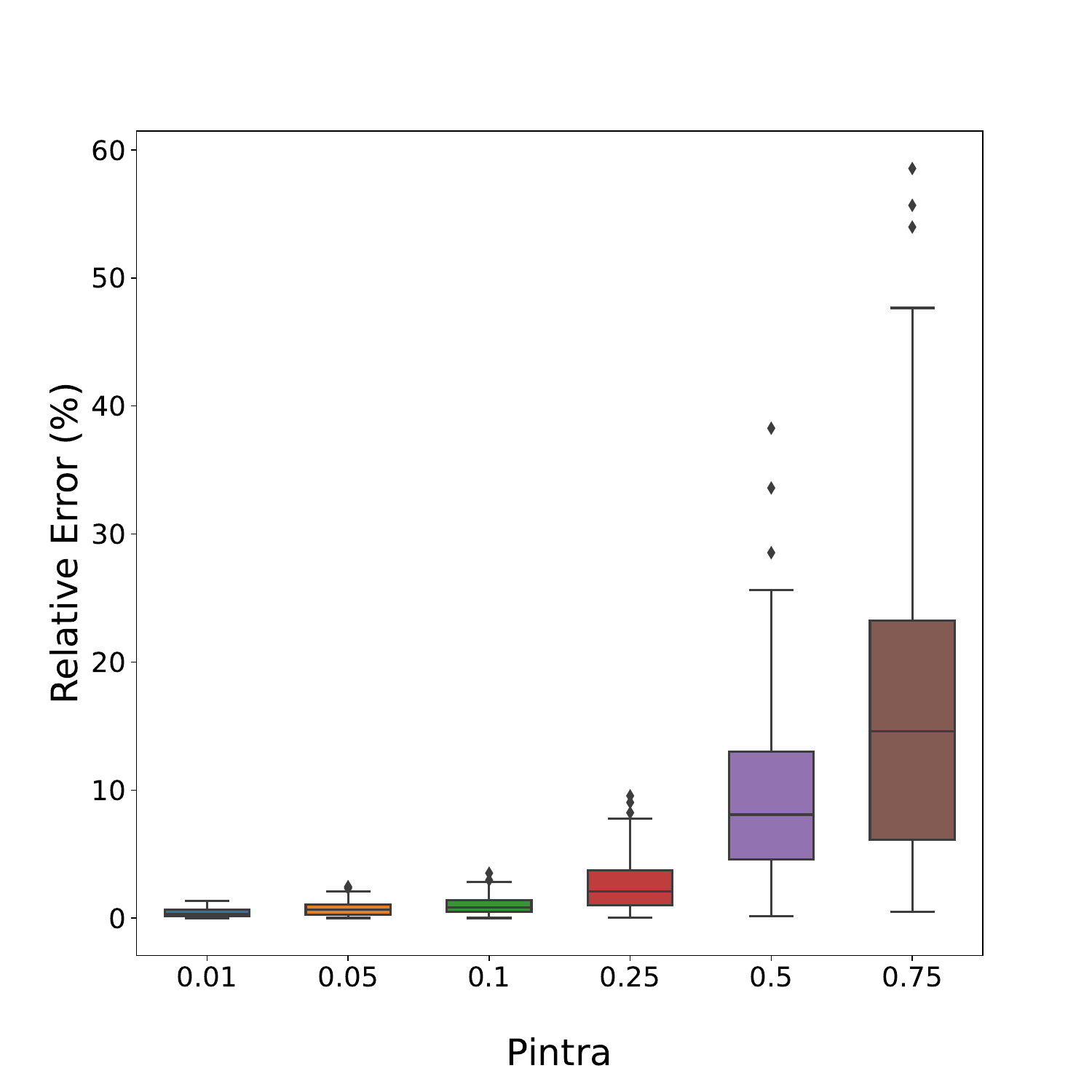}
    \caption{Relative error of SBM for different $P_{intra}$ with $3000$ nodes, $200$ components and $P_{inter} = 0.0001$}
    \label{fig:density}
\end{figure}

\begin{table}[H]
  \centering
  \begin{tabular}{cccc}
    \toprule
    $P_{intra}$ & focal set size & \# edges & MSE \\
    \midrule
     0.01 & 2382 & 655 & 0.052\\
     0.05 & 1788 & 1536 &  0.093 \\
     0.1 & 1349 & 2588 & 0.124 \\
     0.25 & 636 & 5673 & 0.351\\
     0.5 & 271 & 10872 & 1.193\\
     0.75 & 200 & 16157 & 2.120\\
    \bottomrule
  \end{tabular}
  \caption{Mean squared error (MSE) for our Gnn-DML approach on graph generated using a stochastic block model with $3000$ units and $200$ blocks for different values of intra-block tie probabilities, represented as $P_{intra}$.}
  \vspace{0.5cm}
  \label{table:SBM}
\end{table}

\subsection{Generality of GDML Framework: Choice of Nuisance Function Approximator}
For accurate and consistent estimation of nuisance parameters, we leverage the flexible machine learning approach using GNNs. However, our framework can integrate with any graph aggregation tool to estimate propensity scores and outcome models. As the nuisance parameters are functions of both the covariates of an individual unit and those of their social neighbors, their estimation requires aggregating information across the neighborhood. In an effort to demonstrate the generality of our framework, we adopted Network Random Forests (NeRF+) \citep{nerf2024}, which is a family of network-assisted prediction models built upon a generalization of random forests. These models may lack the representational power of GNNs; however, they are interpretable and can be an ideal choice for certain applications.
In our work, we employ the Graph Isomorphism Network (GIN)\cite{xu2018powerful} due to its superior performance over other GNN architectures like GCN~\cite{kipf2016semi}, GAT~\cite{velivckovic2017graph}, and GraphSAGE~\cite{hamilton2017inductive}. GIN's alignment with the representational capabilities of the Weisfeiler-Lehman test \cite{DBLP:conf/aaai/0001RFHLRG19} makes it an ideal choice for effectively capturing the intricate dynamics inherent in social network structures.
Table\ref{table:nerf} presents the result of two variations of our framework combined with GIN and NeRF+ on the Cora dataset with data generative process \ref{dgp1}.
\begin{table}
\vskip 0.15in
\begin{center}
\begin{tabular}{lcccr}
\toprule
    & \textbf{ADE} & \textbf{APE} & \textbf{ATE}\\
\midrule
     GDML + NeRF+ & $1.61_{\scriptsize{{\pm 2.52}}}$ & $87.25_{\scriptsize{{\pm 1057}}}$ & $83.42_{\scriptsize{{\pm 980}}}$\\
     GDML + GIN & $0.16_{\scriptsize{{\pm 0.43}}} $ & $0.21_{\scriptsize{{\pm 0.73}}}$ & $0.40_{\scriptsize{{\pm 1.48}}}$\\
\bottomrule
\end{tabular}
\end{center}
  \caption{The results of application of various graph aggregation tools combined with the core of our framework. MSE over 100 trials is reported.}
  \label{table:nerf}
\end{table}
The performance of GIN in this case is superior; however, the performance of NeRF+ is also close to the ground truth. As mentioned earlier, NeRF+ is interpretable, which may be necessary for some applications.

\end{document}